\pdfoutput=1
\documentclass[11pt]{article}

\usepackage[preprint]{mtsummit25}

\usepackage{copyright}

\usepackage{times}
\usepackage{latexsym}   

\usepackage{float}
\usepackage{ragged2e}
\usepackage{placeins}

\usepackage[utf8]{inputenc}

\usepackage{microtype}

\usepackage{inconsolata}

\usepackage{graphicx}

\usepackage{booktabs} 
\usepackage{array} 
\usepackage{caption} 
\usepackage{listings}
\usepackage{enumitem}
\usepackage{longtable}

\usepackage[avoid-all]{widows-and-orphans}

\newenvironment{compactitemize}
  {\begin{itemize}[itemsep=0pt, parsep=0pt, topsep=0pt]}
  {\end{itemize}}

\title{Translation Analytics for Freelancers:\break
I. Introduction, Data Preparation, Baseline Evaluations}

\author{Yuri Balashov \\
  University of Georgia \\
  Athens, Georgia, USA \\
  \texttt{yuri@uga.edu} \\\And
  Alex Balashov \\
  Evariste Systems, LLC \\
  Athens, Georgia, USA \\
  \texttt{abalashov@evaristesys.com} \\
  \\\And
  Shiho Fukuda Koski \\
  SFK Language Solutions \\
  Rochester, New York, USA \\}

\begin{document}

\maketitle

\begin{abstract}

This is the first in a series of papers exploring the rapidly expanding new opportunities arising from recent progress in language technologies for individual translators and language service providers with modest resources. The advent of advanced neural machine translation systems, large language models, and their integration into workflows via computer-assisted translation tools and translation management systems have reshaped the translation landscape. These advancements enable not only translation but also quality evaluation, error spotting, glossary generation, and adaptation to domain-specific needs, creating new technical opportunities for freelancers. In this series, we aim to empower translators with actionable methods to harness these advancements. Our approach emphasizes Translation Analytics, a suite of evaluation techniques traditionally reserved for large-scale industry applications but now becoming increasingly available for smaller-scale users. This first paper introduces a practical framework for adapting automatic evaluation metrics---such as BLEU, chrF, TER, and COMET---to freelancers' needs. We illustrate the potential of these metrics using a trilingual corpus derived from a real-world project in the medical domain and provide statistical analysis correlating human evaluations with automatic scores. Our findings emphasize the importance of proactive engagement with emerging technologies to not only adapt but thrive in the evolving professional environment.\footnote{ 
Data: \url{https://github.com/YuriBalashov/reeve-corpus}. Code: \url{https://github.com/abalashov/llm-translation-testbed/}.}
\end{abstract}

\section{Introduction}

This is the first in a series of papers exploring the rapidly expanding new opportunities arising from recent progress in language technologies for individual translators and language service providers (LSPs) with modest resources.

\subsection{Background and related work} \label{section:background}

Many translators use MT output in their workflow. In fact, MTPE (machine translation post-editing) has become the default \textit{modus operandi} in the industry \citep{RicoPerez2024} and is seamlessly integrated into computer-assisted translation (CAT) tools and translation management systems (TMS). (For a recent review, see \citealp{moorkens_automating_2025}, Ch. 8.) Most CAT tools can now send real-time queries over the Internet (widely referred to as “API calls”) to any number of generally available neural machine translation (NMT) engines or MT aggregators and present the retrieved translation suggestions to the users for their consideration, alongside translation memory (TM) matches. 

The advent of large language models (LLM) made the work environment of a typical freelancer more complex because, among other things, LLMs can translate, demonstrating performance competitive with that of dedicated NMT engines for some language pairs and domains (\citealp{castilho-etal-2023-online};
\citealp{fernandes-etal-2023-devil};
\citealp{garcia2023unreasonableeffectivenessfewshotlearning}; 
\citealp{hendy2023goodgptmodelsmachine};
\citealp{peng2023makingchatgptmachinetranslation}; 
\citealp{wang-etal-2023-document-level};
\citealp{zhang-etal-2023-machine}; 
\citealp{peters2024didtranslationmodelsrobust}; 
\citealp{li-etal-2024-eliciting}; 
\citealp{li-etal-2024-towards-demonstration};
\citealp{lyu_paradigm_2024};
\citealp{zhu_multilingual_2024}).
Even more importantly, with the right prompting, they can perform increasingly more sophisticated and advanced operations including, but not limited to:
\begin{compactitemize}
    \item Evaluating the quality of translation output, including their own (\citealp{kocmi-federmann-2023-large}; \citealp{lu-etal-2024-error}), with or without reference translations.
    \item Spotting and categorizing translation errors and suggesting corrections (\citealp{berger-etal-2024-prompting}; \citealp{feng2024tearimprovingllmbasedmachine}).
\item Automatic post-editing of raw MT output, including their own (\citealp{raunak-etal-2023-leveraging}; \citealp{Ki2024}; \citealp{alves2024toweropenmultilinguallarge}; \citealp{rei-etal-2024-tower}).
    \item Adapting translation output: 
        \begin{compactitemize}
            \item to the required terminology (\citealp{ghazvininejad2023dictionarybasedphraselevelpromptinglarge}; \citealp{rios2024instructiontunedlargelanguagemodels});
            \item to a given domain (e.g. medical, legal, IT, aerospace engineering, etc.) (\citealp{sia-duh-2023-context}; \citealp{zheng2024finetuninglargelanguagemodels});
            \item to existing translation memories and other project-, client- or domain-specific instructions and reference materials, often outperforming in these respects more traditional approaches earlier implemented in NMT systems (\citealp{moslem-etal-2023-adaptive}; \citealp{moslem2024languagemodellingapproachesadaptive}; \citealp{vieira-etal-2024-much}).
            \end{compactitemize}
            \item Generating mono- and bilingual glossaries of special terms from pairs of source and target documents (\citealp{ding_enhancing_2025}; \citealp{lrag2025})
            \item Improving the quality of translation in low-resource directions (e.g. DE-HI) by following a COT-style (“chain-of-thought,” \citealp{wei2023chainofthoughtpromptingelicitsreasoning}) prompt which explicitly requires them to pivot (“Translate this sentence from DE to EN first; then translate the EN output to HI”); see, in particular, \citealp{jiao2023chatgptgoodtranslatoryes}.
            \item Following, with benefit, a human translation workflow (\citealp{chen-etal-2024-iterative}; \citealp{he-etal-2024-exploring}) by engaging LLMs in a multi-turn interaction involving pre-translation research, drafting, refining, and proofreading \citep{briakou_translating_2024}.
        
\end{compactitemize}

The possibilities in this area are virtually unlimited. Tech giants, larger LSPs, and MT aggregators are losing no time experimenting with these and other approaches in the context of massive localization workflows, with the goal of reducing the role of the proverbial “human expert in the loop” to the very minimum (see, e.g., \citealp{intentomtr2024}; \citealp{aws2025evaluate}; \citealp{rws2025beyond}). CAT and TMS developers are hurrying to incorporate the latest LLM-powered features into their systems (e.g. \citealp{memoq2025agt}: \citealp{bureauworks2025augmented}). New dedicated LLM-based applications are being offered to human translators,\footnote{E.g. \href{https://cotranslatorai.com/}{CotranslatorAI}.} sometimes premised on the assumption that translation memory is a depreciating asset.

\subsection{Our goals in this series of papers}

There is no doubt that these trends will continue to shape the future of translation, human and machine, and will introduce numerous new and unforeseen changes to the fundamental nature of our work. Freelance translators, like everyone else, are adapting to the ongoing changes brought about by the latest developments in AI to the best of their ability. While this adaptation is crucial to the future of the profession, we submit that to get ahead of the curve, a more proactive approach is required.

Linguistic expertise has always been a distinctive mark of excellence in human translation work. However, freelancers are asked to perform other tasks such as sentence alignment, TM clean-up or glossary creation. In our own experience as translators, these tasks are growing in demand, which is consistent with anecdotal evidence from our colleagues and recent industry reports which emphasize “an increasing need for human translators to occupy new roles” \citep{crangasu_adria_how_2025}, such as “AI Content Strategy,” “Big Data Curation,” or “QA Automation” \citep{da_fieno_delucchi_how_2025}. See also \citealp{slator_how_2024, al-batineh_adapting_2024}. 

Freelancers are also increasingly asked to offer their advice on the quality of project- or domain-specific linguistic resources such as TMs or termbases (TB). Use cases include “a company looking to improve its AI translations,” a task that requires “experienced translators to pour through large volumes of the translated text” \citep{crangasu_adria_how_2025}. A request to compare the relative quality of several candidate TMs for a given project is another good example of a task that would benefit from a novel combination of linguistic and technical knowledge. In some cases, pairwise automatic scoring of one TM against another, used as a reference, may be a good first step in the process. We believe that developing new technical skills proactively would make us better prepared for the upcoming challenges. To put it in slogan form, this could make a difference between the “AI is taking our jobs” and “AI is creating new opportunities for us” standpoints pervading much of the current discourse about AI.

Needless to say, many translators already have sophisticated technical capabilities. We think, however, that \emph{Translation Analytics}---an umbrella category we shall use to refer to a variety of methods for the evaluation of the quality of translation-related linguistic assets---have not been deployed by freelancers to its full capacity. In fact, for most of them, ‘Translation Analytics’ may be synonymous with pre-translation analysis performed by CAT tools to generate the statistics for fuzzy TM matches at the start of a new project---for pricing, time planning, and other business purposes. Translation Analytics, however, are much broader in scope. We think of them as including, but not limited to:
\begin{compactitemize}
    \item Human evaluation methods ranging from linear scoring to Multidimensional Quality Metrics  (\citealp{burchardt-2013-multidimensional}; \citealp{freitag-etal-2021-experts}; \citealp{knowles_calibration_2024}; \citealp{lommel-etal-2024-multi}).
    \item Automatic evaluation metrics, such as BLEU \citep{papineni-etal-2002-bleu}, chrF \citep{popovic-2015-chrf}, TER \citep{snover-etal-2006-study}, and COMET \citep{rei-etal-2020-comet}.
    \item Any number of \textit{ad hoc} tools and methods for statistical analysis and quality estimation that may be developed for a given project and tailored to its specific demands.
\end{compactitemize}

Our main goal in this series of papers is to explore the full potential of Translation Analytics in the context of a typical freelancer workflow. We aim to empower fellow translators with new methods that would allow them to add value to their services at the time of big changes and to gain control of the processes that usually happen “under the hood.” We also hope this will stimulate developers of CAT/TM tools and TMS systems to incorporate some of the analytic methods we describe in this series of papers into their products.

In the end, freelancers  should be able to implement many of the sophisticated operations mentioned in Section \ref{section:background} above, in their local translation environment, with practical, theoretical, and strategic benefits. Instead of contributing the last, indispensable but increasingly small, bit of human expertise to the proverbial “loop” the translator can get back into the driver’s seat by learning a small number of new technical skills.

\subsection{Our goals in this first article}
\label{subsection:goals_first_article}

In the first article in this series we focus on adapting automatic evaluation metrics to the needs and work environment of individual translators and smaller LSPs who may want to take their technical capabilities to the next level. 

Automatic evaluation of MT quality has been a prominent focus in the industry for years. Traditional metrics like BLEU \citep{papineni-etal-2002-bleu}, chrF \citep{popovic-2015-chrf}, and TER \citep{snover-etal-2006-study} assess the output of MT systems by comparing it to reference translations, ideally created by skilled human translators. These comparisons rely on word or character-level string matching. Newer metrics such as COMET \citep{rei-etal-2020-comet}, BLEURT \citep{sellam-etal-2020-bleurt}, and BERTScore \citep{Zhang2020BERTScore} evaluate translations within the semantic space of neural networks. This approach is less reliant on specific word choices and instead prioritizes the underlying linguistic meaning.

The correlation between automatic metrics and human evaluation remains a topic of debates (for a recent overview of these debates, see \citealp{moorkens_automating_2025}, Ch. 5), yet these metrics are essential in MT research and development. They enable developers to quickly compare model outputs after numerous adjustments to determine whether a particular change improves quality. Additionally, automatic metrics can monitor the training of NMT models by calculating, for instance, a BLEU score on a reserved test set after each iteration. Training can be stopped when no further improvement is observed.

Historically, automatic metrics were both technically complex and irrelevant to human translators, who depended on their linguistic expertise and manual analysis. However, with the seamless integration of MT engines into CAT tools, the vast availability of bilingual data at translators' fingertips, and recent advancements in generative AI, the landscape is evolving rapidly. Many translators now incorporate MT into their workflows and often need to choose among multiple MT engines for specialized projects, sometimes spanning tens of thousands of words. Translators frequently possess valuable bilingual resources, such as TMs and TBs from similar projects, which allow them to evaluate MT engine outputs in minutes using automatic metrics. Free online tools designed for users without programming expertise facilitate this process.\footnote {And many other processes. Thanks to free online toolkits such as \href{https://github.com/adaptNMT}{adaptNMT} \citep{lankford_adaptnmt_2023}, anyone can now build, train, fine-tune, and evaluate an NMT system more or less from scratch!} One such tool, MATEO (MAchine Translation Evaluation Online) \citep{vanroy-etal-2023-mateo}, is used in our work.

To illustrate the power and practical value of such methods for individual translators, we need high-quality data---parallel documents in two or more languages. While most of industrial-scale translation quality evaluation research is based on the datasets made available on Workshops on Machine Translation (WMT) benchmarks  \citep{kocmi-etal-2024-findings} and other shared task repositories, we take our data from a recent real-life translation project completed in summer 2024 for a non-government organization, as described below.

Our main contributions detailed in this paper are as follows:
\begin{compactitemize}
    \item We present, with the client’s permission, a \textbf{trilingual corpus of over 4.5K sentences in English, Russian, and Japanese in the medical domain (the Christopher \& Dana Reeve Foundation Trilingual Corpus, RFTC)}, resulting from a recent human translation project completed by YB (EN-RU) and SFK (EN-JA) who are certified by the American Translators  Association in their respective language pairs. We hope this corpus will be used for non-commercial research purposes by others and that it will grow both in coverage and language varieties. 
    \item We use this corpus to develop and implement a relatively simple \textbf{approach to translation quality evaluation} which can be adapted by technically oriented translators and LSPs with modest resources to assess the quality of translation output from traditional NMT engines and LLMs in an informed way.
    \item We report the \textbf{BLEU}, \textbf{chrF2}, \textbf{TER}, and \textbf{COMET} scores for the translation outputs (EN-RU and EN-JA) for a slightly smaller but more challenging version of our corpus (about 3.5K English sentences) from \textbf{three popular NMT engines} (labeled MT1–MT3) and \textbf{three popular LLM models} (LLM1–LLM3), using our professional human translations for reference (see Appendix \ref{appendix:NMTLLMsUsed} for details).

    \item We adopt a simple linear 0.0–4.0 scale \textbf{modeled after academic grading} to perform preliminary \textbf{human evaluation} of \textbf{540} MT- and LLM-produced translations in each of our target languages (i.e. \textbf{1080} sentences overall).
    \item We report and discuss the results of our \textbf{preliminary statistical analysis} in order to determine:
        \begin{compactitemize}
            \item whether the automatic scores computed for smaller non-overlapping parts of our source document (229, 1143, and 2183 sentences) correlate with each other;
            \item whether sentence-level COMET scores for select segments for each output correlate with the human grades for them.
        \end{compactitemize}
\end{compactitemize}

Two of the authors (YB and SFK) are ATA-certified professional translators with little or no programming experience or skills. Our perspective, therefore, fits the goals of this use case study. We should add that while we could, in principle, meet our coding needs by asking LLMs to write simple programs for our operations, coding with LLMs can be a haphazard process; the output can be very good and correctly focused on the problem, or can be mediocre and not especially applicable to what one is trying to accomplish. Stitching it all together for the purpose of a systematic study is still a task that benefits from a great deal of human expertise. Our experiments would be far from complete at this point without the tremendous help from a professional programmer on our team (AB) who took care of all the LLM-related operations, API call parallelization, streamlining, and more, as described in Section \ref{section:TranslationOutputs} below.

We adapt our discussion throughout the paper to the specific needs of individual translators and smaller LSPs. While the size of our corpus is small by MT industry standards, it is quite large for a single human translation project, and it generates statistically significant evaluation data. Furthermore, since our corpus is unlikely to have been seen and used for training or fine-tuning by generic NMT engines and popular LLMs at the time of conducting our baseline experiments, it adds new evidence for the ongoing debate about the quality and reliability of automatic quality metrics. 

We believe that getting under the hood of translation quality evaluation is very important for freelancers and smaller LSPs at the time when traditional workflow models are being replaced by increasingly more sophisticated tasks requiring new technical expertise and willingness to learn more advanced methods. We submit that equipping individual translators with the additional technical capabilities described in this series of papers will help them adapt their toolkits to the rapidly changing work demands and new challenges brought about by the rocket speed development of language technologies.

The plan for the paper is as follows. Section \ref{section:TrilingualCorpus} presents the Reeve Foundation Trilingual Corpus, complete with our reference translations. Section \ref{section:TranslationOutputs} describes how we obtained MT and LLM translation outputs for our source documents. In Section \ref{section:AutomaticQualityEvaluation} we report and discuss the automatic metric scores for the entire corpus. In Section \ref{section:SampleSize} we investigate pairwise correlations among the scores for three smaller parts of the corpus. In Section \ref{section:ManualEvaluation} we develop our approach to human evaluation of select MT and LLM output, present its results, and discuss their statistical significance. In Section \ref{section:Limitations} we note the limitations of our study and outline plans for future work. Section \ref{section:Conclusions} summarizes our findings and conclusions.

\section{The Christopher \& Dana Reeve Foundation Trilingual Corpus}\label{section:TrilingualCorpus}

We illustrate our Translation Analytics methods with the resources from a large translation-editing-proofreading project completed in summer 2024 for the Christopher \& Dana Reeve Foundation.{\footnote{The authors thank the \href{http://www.christopherreeve.org/}{Christopher \& Dana Reeve Foundation} for a kind permission to use their linguistic resources in this work.}} Specifically, the Foundation’s \href{https://www.christopherreeve.org/todays-care/living-with-paralysis/free-resources-and-downloads/paralysis-resource-guide/}{Paralysis Resource Guide} is “a free comprehensive 392-page book designed to empower individuals living with and impacted by paralysis to lead healthy and fulfilling lives.” A shorter (80K words) international edition of the Guide was recently translated into several languages. The Guide (referred to below as ‘PRG’) is a coherent structured document divided into chapters and sections, complete with a descriptive glossary of about 200 technical terms. The translation project (EN-RU and EN-JA) came in the form of IDML (InDesign Markup Language) files for separate chapters. The PDF layouts of the EN, RU, and JA versions of PRG are included in our corpus for reference.

As the first step in data preparation, we took the versions of our TMs which preserve the order of source sentences in the original full document. We removed IDML and other tags from the TMs, discarded repetitions, and produced a spreadsheet that combined the source text (EN) and our reference translations (RU, JA). Next, we performed additional cleanup operations to remove:

\begin{compactitemize}
    \item leading and trailing spaces;
    \item bullets and other special characters at the beginning of segments;
    \item segments with only or mostly numbers;
    \item segments with only or mostly URLs;
    \item segments with only or mostly address lines or phone numbers.
\end{compactitemize}

The resulting Excel file \texttt{1-10\_en\-ru-ja\_long.xlsx} contains 4528 segments supplied with stable ID numbers (Column A), which are used in all our experiments.

To make the translation task more challenging for MT engines and LLMs, we also decided to remove segments shorter than 6 source words from our set and generated “short” versions of the data (\texttt{1-10\_en-ru-ja\_short.xlsx}, etc.). The source sentence length (\texttt{Len}) is calculated in Column F.

An additional minor reduction was necessitated by the limitations MATEO imposes on the input file size ($\leq$ 1MB) for evaluation (Section \ref{section:AutomaticQualityEvaluation} below). To preserve the natural order of the segments, we met this requirement by removing the last two parts of PRG (“Glossary” and “Back Cover”), which brought the segment count down to 3555 (\texttt{1-8\_en-ru-ja\_short.xlsx}). The resulting Excel document was used to prepare tri- and monolingual Unicode text files for our experiments.

The materials referenced above comprise the Christopher \& Dana Reeve Foundation Trilingual Corpus (alternatively, the Reeve Foundation Trilingual Corpus, RFTC), complete with the PDF layouts. Additional corpus details can be found in Appendix \ref{appendix:TrilingualCorpus}. With the client’s permission, we make the corpus described here available for non-commercial/academic use.

\section{Translation Outputs}\label{section:TranslationOutputs}

In this section we describe how we obtained MT and LLM translation outputs for our corpus.

\subsection{Technical notes on MT output}

To preserve data confidentiality, we used the “Pro” versions of three popular NMT engines (labeled MT1, MT2, and MT3) to translate the entire \texttt{1-10\_en\_short.txt} document (3896 segments, one per line). The process was implemented as “pre-translation” in memoQ for MT1 and MT3, and was performed directly for MT2. We tracked the run-times for these operations (Table \ref{tab:mt_llm_runtimes} below).

\subsection{Technical notes on LLM output}\label{subsection:tech-notes-LLM}

Translation with LLMs was more complicated.  API calls over the Internet must be used to interact with the major LLMs because the latter offer both the use of the model, and the significant parallel computing resources required to run it, as an integrated, metered “cloud” service. We used the Python programming language and the Python SDKs provided by major LLM vendors. We used paid subscription accounts for all LLM calls, with maximum data security/privacy settings allowed for these accounts. 

\subsection{Bulk processing and LLMs}

There are, in principle, a number of ways to feed a large list of sentences to major LLMs. Some of them, for example, offer API constructs for batch processing, specifically intended for non-time-critical bulk tasks. In this approach, large data sets are uploaded for the LLM provider's backend to churn through on a best-effort basis. To limit scope creep and eliminate variation in how we used different LLMs, we did not explore this option. It is also possible to submit multiple sentences with every request; this we did try, but we found the formatting characteristics of the resulting output to be too inconsistent for automatic evaluation. Therefore, the only method we evaluated was sentence-by-sentence, with one sentence per request. 

It is worth taking a moment to reflect on the fact that this sentence-by-sentence approach is relatively naive, in a sense, even if it also eliminates some confounding factors. In contrast to the contextual environment of an ongoing ChatGPT conversation, in which the model keeps a running context window where prior prompts and responses reside, every one of our API requests instantiated a \textit{de novo} context that was not informed by prior state. We did not attempt to evaluate the impact of context windows upon translation quality for two reasons: (1) the additional variables introduced would be unwieldy for the modest ambitions of this paper, and (2) some \textit{ad hoc} experimentation did not suggest that there was much, if anything, to be gained in translation \textit{quality} this way, and therefore it did not seem a propitious avenue for our specific aims. Still, this may be worth exploring in future research.

\subsection{Prompt specificity}

We found that brief and broad requests are not rewarded with as much consistency as long and specific ones. For example, when commanded to \textbf{\texttt{\small "translate the following sentence to Russian:\_\_\_\_\_\_\_\_"}} major LLMs would, for the most part, return the translated sentence and nothing else. However, every once in a while, the resulting sentence would contain additional verbiage: \textbf{\texttt{\small "Here is the following sentence in Russian: \_\_\_\_\_\_\_"}}.

With a more laborious prompt, which spelled out some examples of extraneous contributions unrelated to the translation of the sentence, this effect could be mostly, but not entirely obviated:

\textbf{\texttt{\small "You are an expert translator, translating for an expert audience. Please do not provide any annotations, explanations or transliterations in your translation. Please translate the following sentence to Russian (Japanese): \_\_\_\_\_\_\_\_"}}

Rarely, extraneous output would still appear, although the prompt was highly effective at reducing the incidence of it. (We did not specifically attempt to measure the incidence.) This is a salient consideration for any endeavor that relies on low-touch bulk translation by LLMs.

\subsection{Temperature and determinism}\label{subsection:temperature}

It is well known that LLMs' output is not 100\% deterministic. All LLM providers offer an API call parameter called “temperature” ($T$) which regulates the degree of acceptable stochastic variance in responses; higher temperatures allow more randomness, and lower values less. We set $T=0.0$ in all of our requests across the board, but occasional variation in responses to identical prompts, while rare, was still present. 

\subsection{“Buggy” prompts}\label{subsection:buggy}

Upon completing all the operations with LLM and collecting all the outputs, we discovered that our optimized prompting routine concatenated the language name (i.e., ‘Russian’ or ‘Japanese’) to the prompt prefix twice:\label{buggy}

\textbf{\texttt{\small "You are an expert translator, translating for an expert audience. Please do not provide any annotations, explanations or transliterations in your translation. Please translate the following sentence to \textcolor{blue}{Russian (Japanese):} \textcolor{red}{Russian (Japanese):} \_\_\_\_\_\_\_\_"}}

Given the length of the prompt, we hypothesized that this did not have a significant impact on the output. But we decided to perform a safety check comparing the LLM outputs for a shorter part of PRG (\texttt{5.en\_short}, 229 segments) with the above prompt (which we used in our experiments) as well as  with the corrected prompt:

\textbf{\texttt{\small "You are an expert translator, translating for an expert audience. Please do not provide any annotations, explanations or transliterations in your translation. Please translate the following sentence to \textcolor{blue}{Russian (Japanese):} \_\_\_\_\_\_\_\_"}}

We generated two outputs with the “bug-free” prompt to see if the differences between them due to the usual sampling (even with $T=0.0$) in LLMs are significantly smaller than the differences between each of them and the output for the “buggy” prompt. The results reported in Appendix \ref{section:RuntimeDetails} suggest that the answer is No. In terms of automatic scores, the differences among the three outputs are marginal and statistically insignificant, both for EN-RU and EN-JA. Interestingly, the “buggy” prompt actually did slightly better!	

Tempting as it was to call it a feature not a bug, our safety check leads us to categorize it as insignificant and discardable statistical noise. We add this to the growing list of observations of rather unpredictable sensitivity of LLMs’ output to the fine details of the prompts in some cases, and their surprising robustness to prompt changes in other cases. We further hypothesize that LLMs’ insensitivity to the potentially misleading second occurrence of ‘Russian’ (or ‘Japanese’) in the “buggy” prompt may have to do with (i) their default preference for English; and/or (ii) their ability to identify the language of the string that actually follows ‘:’; and/or (iii) the fact that transformer-based neural networks, unlike the older LSTM- and GRU-based architectures, compute the attention scores between all pairs of tokens in the entire input directly and in parallel, rather than consecutively, so the fact that the second occurrence of the language name (‘Russian’ or ‘Japanese’) immediately precedes the source sentence does not make the former more important than the other preceding tokens.

We release all translation outputs from the systems we tested in the form of a single Excel file named \texttt{\small 1-10\_en-ru-ja\_short\_MT-LLM-outputs.xlsx}, where Column A contains the segment IDs, Column F the source segment length (in words), and the other columns are labeled with the target language and the system which generated the output.

\section{Automatic Quality Evaluation}
\label{section:AutomaticQualityEvaluation}

In this section we report and discuss the automatic metric scores for the entire corpus.

As already noted in Section \ref{section:TrilingualCorpus} above, we had to reduce the length of our corpus by about 9\% to 3555 segments to meet the file size requirements of MATEO \citep{vanroy-etal-2023-mateo}, the tool we utilized to calculate the BLEU, chrF2, TER, and COMET scores for our outputs. We provide additional details in Appendix \ref{section:auto-quality-details}.

The evaluation scores for \texttt{1-8\_en} are represented in Table \ref{tab:side_by_side} and Figure \ref{fig:full_width_image} below. Consistently with other reports, the string-based scores for EN-JA are lower than for EN-RU. We note, however, that the COMET scores are neck-to-neck; in fact, slightly higher for EN-JA for all LLM outputs and MT3. All the score differences are statistically significant.

Since the distinction between the linguistic concepts of character and word is blurred in Japanese, questions may be raised about the separate significance of chrF for translation directions involving this language. We do not have a considered view on this. But we calculated pairwise Pearson correlation values for BLEU-chrF2, BLEU-TER, and BLEU-COMET between the scores for our six systems for both language pairs (Table \ref{tab:pearson_correlations}). We note high correlations between BLEU and chrF2, and between BLEU and TER for both language pairs, a somewhat lower but still solid correlation between BLUE and COMET for EN-RU, and the lack of correlation between BLUE and COMET for EN-JA. Along with COMET’s neck-to-neck results for both language pairs, this underscores the importance of neural-based metrics.

In our experiments, performed on a 13th Gen Intel(R) Core(TM) i9-13900KF 3.00 GHz 64.0 GB PC, MATEO took roughly 20 minutes to compute the four scores for a single output against a reference; for JA it took slightly longer than for RU. Of these 20 minutes, roughly 16 minutes go into computing the COMET scores, 2 minutes into TER, and 2 minutes into bootstrap resampling at the very end. The calculation of BLUE and chrF2 is very fast. In light of the above-noted considerations, the time spent on computing the COMET scores is the time well spent. Users should be aware of this.

\section{What Sample Size is Needed for Reliable Automatic Quality Evaluation?}\label{section:SampleSize}

Another important question that may arise for freelancers inclined to use automatic evaluation of MT/LLM outputs in choosing the best system for a new project is the minimal size of a sample required to make a reliable decision. A freelancer may have a good TM from a previous project in the same domain or for the same client that could be used for reference. Alternatively, a freelancer may complete a representative part of a new project and decide to add the best-performing MT and LLM-based system to their workflow going forward. One can imagine similar scenarios. Such deliberations should, of course, take into account typological differences between target languages which may affect the automatic scores for string- and neural-based metrics differently. In all cases of this sort, the size of the sample to be used for MT/LLM translation quality evaluation must be statistically significant. What is the minimal size that meets this requirement?

To approach this question empirically we generated additional sets of automatic MT/LLM evaluation scores for the outputs from three distinct parts of our corpus, \texttt{229\_en} (229 segments, identical to \texttt{5\_en\_short}), \texttt{1143\_en} (1143 segments, identical to \texttt{3\_en\_short}), and \texttt{2183\_en} (2183 segments comprising the rest of \texttt{1-8\_en\_short}) to see how well they correlate with each other. Tables \ref{tab:evaluation_scores} and \ref{tab:evaluation_scores_japanese} in Appendix \ref{section:eval-scores-diff-sizes} feature the four sets of scores, including those for \texttt{1-8\_en\_short} (3555 = 228 + 1143 + 2183 segments). 

The lack of overlap among \texttt{229\_en}, \texttt{1143\_en}, and \texttt{2183\_en} (cumulatively comprising the entire \texttt{1-8\_en\_short} document), which is evidenced in our memoQ analysis (Table \ref{tab:memoQanalysis}) makes them suitable for correlation analysis, as does their thematic coherence: all three originate in a single narrow-domain document. Tables \ref{tab:correlations} and \ref{tab:correlations_enja} (Appendix \ref{section:pearson-three}) represent the Pearson correlation values $r$ along with their $p$-values for three pairs of evaluation scores sets corresponding to \texttt{229\_en}, \texttt{1143\_en}, and \texttt{2183\_en}.

We observe that the correlations are very strong in all cases, across all the metrics. We are thus led to conclude that computing automatic scores for a small part of our document (229/3555 = 6.4\%) would give us a good sense of the relative performance of several MT/LLM systems. However, this approach has its limitations. See Appendix \ref{section:pearson-three} where we also provide additional details regarding the use of statistical methods for freelancers and discuss the prospects for future work.

\begin{table*}[htbp]
\centering
\begin{minipage}{0.48\textwidth}
    \centering
    \begin{tabular}{lcccc}
        \toprule
        & \textbf{COMET} & \textbf{BLEU} & \textbf{chrF2} & \textbf{TER} \\ \midrule
        MT1  & 88.1          & 41.1          & 64.4           & 43.1         \\
        MT2  & \textbf{90.8} & \textbf{57.2} & \textbf{74.2}  & \textbf{31.1} \\
        MT3  & 90.2          & 45.4          & 67.4           & 40.0         \\
        LLM1 & 88.8          & 38.4          & 63.5           & 45.4         \\
        LLM2 & 89.3          & 37.1          & 63.0           & 46.2         \\
        LLM3 & 88.6          & 33.2          & 60.1           & 50.1         \\ 
        \bottomrule
    \end{tabular}
    \caption*{\textbf{English-Russian}}
\end{minipage}%
\hfill
\begin{minipage}{0.48\textwidth}
    \centering
    \begin{tabular}{lcccc}
        \toprule
        & \textbf{COMET} & \textbf{BLEU} & \textbf{chrF2} & \textbf{TER} \\ \midrule
        MT1  & 88.1          & 31.1          & 39.5           & 55.3         \\
        MT2  & 89.7          & \textbf{38.6} & \textbf{46.0}  & \textbf{47.5} \\
        MT3  & \textbf{90.6} & 36.8          & 44.1           & 49.7         \\
        LLM1 & 89.5          & 31.9          & 38.6           & 53.0         \\
        LLM2 & 90.1          & 30.2          & 37.6           & 53.9         \\
        LLM3 & 89.5          & 28.9          & 36.3           & 55.2         \\ 
        \bottomrule
    \end{tabular}
    \caption*{\textbf{English-Japanese}}
\end{minipage}
\caption{Evaluation metric scores for MT and LLM models for English-Russian and English-Japanese translations for \texttt{1-8\_en\_short}.}
\label{tab:side_by_side}
\end{table*}

\begin{figure*}[htbp]
\centering
\includegraphics[width=\linewidth]{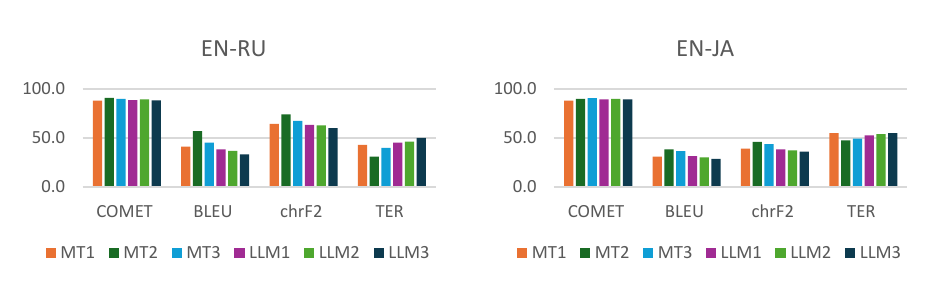}
\caption{Visualization of MATEO-generated metric scores for EN-RU and EN-JA translations, broken down by MT engine and LLM, for \texttt{1-8\_en\_short}.}
\label{fig:full_width_image}
\end{figure*}

\section{Manual Evaluation of Select Translation Outputs}\label{section:ManualEvaluation}

We selected 180 and 360 MT- and LLM-translated sentences from the outputs for \texttt{5\_en\_short} and \texttt{3\_en\_short} respectively for each of our language pairs (i.e. 1080 segments overall) to perform manual evaluation of their quality with a simple linear scale in order to estimate whether the outputs’ sentence-level COMET scores correlate with our “human grades” for them. Below we outline our selection process, the evaluation scale, and the results.

\subsection{Segment selection}\label{subsection:SegmentSelection}

We ranked MT- and LLM-generated translations by their sentence-level COMET scores and selected 10 highest-scoring segments, 10 intermediate-scoring segments, based on their median ranks, and 10 lowest-scoring segments from the outputs for \texttt{5\_en\_short}. We doubled these numbers (20-20-20) for \texttt{3\_en\_short}.

\subsection{Human grading}\label{subsection:HumanGrading}

To assign “human grades” to the selected translations we adopted a linear 0.0–4.0 scale modeled after academic grading (Table \ref{tab:grading_scale} in Appendix \ref{section:manual-grading-scale}). Two of the co-authors who have extensive academic teaching experience found this approach intuitive and efficient: it is easy for them to imagine they are grading student work. Along with the letter/numeric grades, we supplied brief notes for each graded translation highlighting 1–2 most serious issues from the following list: Accuracy; Clarity; Consistency; Fluency; Grammar (including spelling, typography, and syntax); Register; Style; Terminology; Tone. To minimize our bias in grading, we sorted these segments by their ID numbers rather than by their COMET scores.

While we are fully aware of the multiple limitations of this approach, our primary goal in this first round of baseline evaluations was to develop and offer to fellow translators a potentially fruitful method that would allow them to see whether automatic scores correlate with their human judgment in their particular use case.

\subsection{Are automatic evaluation scores correlated with human grades?}\label{subsection:Correlation}

We calculated Pearson ($r$) and Spearman ($\rho$) correlation coefficients between sentence-level COMET scores and our numeric grades for the 10-10-10 and 20-20-20 selections from each translation output (Tables \ref{tab:pearson_spearman} and \ref{tab:pearson_spearman2} in Appendix \ref{section:human-comet-correlation}) for each language pair. Most of the $r$ and $\rho$ values suggest moderate to strong correlation; but the variation is rather wide, between MT/LLM outputs, language pairs, and sample sizes. Some of the variation may an artifact of our somewhat impressionistic and non-rigorous grading and/or the sampling method. These may be adjusted depending on the available human resources. But calculating sentence-level correlations is a very natural and easy strategy to pursue in all cases where “human grades” of select outputs are available.

We release all the selected sentences along with their COMET scores and our grades and comments in the form of two Excel files: \texttt{3\_en-ru-ja\_short\_comet\_grades.xlsx} and \texttt{5\_en-ru-ja\_short\_comet\_grades.xlsx}.

\section{Limitations of Our Study and Future  Plans}\label{section:Limitations}

\textbf{Translation directions}. We had an opportunity to experiment with two interestingly different language pairs because we ourselves produced the translations for them and the client gave us permission to use them. We hope that other language directions and document families will be added to our corpus in the future.

\noindent \textbf{Automatic metrics}. We limited our choice of them to BLEU, chrF2, TER, and COMET, to maximize efficiency and ease of use. More advanced users should consider other metrics and consult the current best practices \cite{kocmi-etal-2024-findings}.

\noindent \textbf{Correlation experiments with sample sizes} reported in Section \ref{section:SampleSize} need to be complemented with power analysis to determine the minimal size of the statistically significant sample. Ideally, future experiments should also include other contrasting pairs---from different domains, registers etc.

\noindent \textbf{Human evaluation} requires independent raters and a uniform blinding and randomization protocol. While extensive, our reported results must be taken with a grain of salt. We do believe they serve as a proof of concept.

\noindent \textbf{In future experiments} with our corpus (RFTC) we want to explore the potential of various dedicated systems and LLMs for (i) extracting a bilingual glossary from a set of parallel sentences, and (ii) using a glossary thus obtained to improve the quality of translation in the context of the freelancer’s workflow.

\noindent \textbf{Other ideas} are briefly described in Section \ref{section:background} above. We may pursue some of them and invite fellow translators and other interested parties to join us in this effort.

\section{Conclusions}\label{section:Conclusions}

This study demonstrates the potential of Translation Analytics to help freelance translators and smaller language service providers (LSPs) thrive in a rapidly evolving industry. By adapting evaluation metrics such as BLEU, chrF, TER, and COMET to individual workflows, we provide methods for assessing MT and LLM outputs with rigor and precision.
The findings underscore several critical insights:

\noindent \textbf{Utility of automatic evaluation metrics}. Automatic metrics, particularly COMET, consistently align with human assessments, reinforcing their value as robust tools for translation quality evaluation. Translators can confidently leverage these metrics to make informed decisions about incorporating MT and LLM systems into their workflows.

\noindent \textbf{Efficiency of sample-based evaluation}. Even small, strategically selected samples of documents can yield statistically reliable insights into the relative performance of different translation systems. This approach enables resource-efficient evaluation for freelancers working on large-scale projects.

\noindent \textbf{Integration of human judgment}. While automatic metrics are helpful, the integration of human evaluation, anchored in linguistic expertise, remains critical. Our experiments validate the complementary roles of human judgment and automated tools in achieving nuanced and accurate quality assessments.

\noindent \textbf{Empowering freelancers}. By demystifying technical methods and tools, we equip translators with the confidence and skills to engage proactively with advanced language technologies. We hope this will help them move beyond being mere participants in the workflow to assuming leadership in optimizing and innovating translation practices. We offer one concrete entry point, with examples of expanded capabilities, in Appendix \ref{section:apis-applied-technical-avenues-freelancers}.

Future work will focus on expanding the corpus to include additional language pairs, domains, and registers to further validate and refine our methods. Moreover, exploring advanced techniques such as glossary extraction, domain-specific adaptation, reference-free quality estimation, automatic post-editing, and more sophisticated multi-step operations using LLMs represents promising avenues for enhancing translation quality and efficiency.

As the landscape of translation continues to evolve, it is imperative for freelance translators and smaller LSPs to embrace new tools and methodologies. By doing so, they can not only adapt to the changes but also seize the opportunities presented by advancements in language technology. This proactive approach will ensure that translators remain at the forefront of a profession that is as dynamic as it is indispensable.

\section*{Author Contributions}

YB developed the initial plan, prepared the data, ran the evaluation experiments, and wrote most of the content including literature review and bibliography, but excluding Sections \ref{subsection:tech-notes-LLM}--\ref{subsection:temperature} and Appendix \ref{section:apis-applied-technical-avenues-freelancers}, which were contributed by AB, who took care of all our programming needs. YB and  and SFK are ATA-certified translators who worked with their partners in summer 2024 on translating the International Edition of the Reeve Foundation's Paralysis Resource Guide to Russian and Japanese. They performed manual evaluation of the 1080 selected MT- and LLM-generated segment translations as described in Section \ref{section:ManualEvaluation}. They also provided additional notes on the RFTC corpus in Appendix \ref{appendix:TrilingualCorpus}. SFK curated the Japanese portion of the data.

\section*{Sustainability Statement}

Our experiments performed on personal computers did not involve training of neural models. Computing COMET scores and querying LLMs for translation were the longest operations. We report the runtimes for them in Section \ref{section:AutomaticQualityEvaluation} and Table \ref{tab:mt_llm_runtimes}.

We used one of the recommended algorithms\footnote{\url{https://calculator.green-algorithms.org/}} to estimate a carbon impact of our computations according to \cite{lannelongue_green_2021}. A brief report is included it Appendix \ref{section:carbon-imprint}.

\section*{Acknowledgments}

YB's work is supported by the NSF grant No. \href{https://www.nsf.gov/awardsearch/showAward?AWD_ID=2336713}{SES-233671}. We are grateful to  Bran Vanroy for helpful comments and clarifications on MATEO. We thank the reviewers for their very helpful comments. We reiterate our thanks to the Christopher \& Dana Reeve Foundation for the permission to use their linguistic resources in our experiments.

\bibliography{mtsummit25}

\appendix

\section{NMT Engines and LLMs Used in our Experiments}\label{appendix:NMTLLMsUsed}

MT1 = ModernMT Professional

\url{https://www.modernmt.com/translate}

\noindent MT2 = DeepL Translator Pro

\url{https://www.deepl.com/en/translator}

\noindent MT3 = Google MT (Cloud Basic) 

\url{https://translate.google.com}

\noindent LLM1 = \texttt{GPT-4o}

\url{https://platform.openai.com/docs/models/gpt-4o}

\noindent LLM2 = \texttt{Claude 3.5 Sonnet}

\url{https://console.anthropic.com}

\noindent LLM3 = \texttt{Gemini 1.5 Pro}

\url{https://ai.google.dev/gemini-ap}

We selected these engines and models at the time of conducting our baseline experiments (November 2024 – January 2025) based on a balance of the following considerations:

\begin{compactitemize}
    \item their popularity among freelance translators and LSPs with limited resources;
    \item their subscription and per-token costs;
    \item their existing integration into CAT tools.
\end{compactitemize}

There are numerous other options available, including new NMT systems and the latest LLMs, and we plan to explore some of them in the future.
We also believe that at the time of our initial experiments reported here, popular LLMs and NMT systems have not seen our trilingual data and, hence, could not have used it for re-training or fine-tuning. Now that this data is available, it might be of some interest to see if our chosen models’ performance has changed \citep{kocyigit_overestimation_2025}.

\section{The Reeve Foundation Trilingual Corpus: Additional Details}\label{appendix:TrilingualCorpus}

The source document statistics for our corpus are compiled in Table \ref{tab:source_document_statistics} below. Table \ref{tab:memoQanalysis} presents memoQ analyses of both inputs, “long” and “short.” Table \ref{tab:prg_table} provides further details of the corpus.

Although translations of the Reeve Foundation International Edition of the Paralysis Resource Guide (PRG) are intended to be generally available, their main target is the US population for whom English is a second language.

In the Russian translation of PRG, organization and program names and most of their acronyms were translated on their first occurrence followed by the English original and acronym in parentheses. In subsequent occurrences in the same section of the document, only translations or translated acronyms (where available) were used. Exceptions include acronyms such as ‘FDC’ and brand names of companies and their products, such as ‘Pfizer’ and ‘Tobii Dynavox’, which are kept in English. The brand medication names were translated or transliterated followed by their original English names on their first appearance. Only translations were used on subsequence occurrences. Number notation generally follows Russian conventions, i.e. ‘33,000’ $\rightarrow$ ‘33 000’; ‘6.79’ $\rightarrow$ ‘6,79’; etc.

The Japanese translation of PRG generally adheres to the notation guidelines outlined in \href{https://www.jtf.jp/pdf/jtf_style_guide_e.pdf}{JTF Style Guide for Translators Working into Japanese}. A polite and neutral style using the \emph{desu/masu} form was applied, and the honorific suffix \emph{san} was added after the names of individuals outside the Reeve Foundation. All personal names were transliterated. For medical terms, the original English term and its Japanese translation were juxtaposed in the headings of each section, separated by a slash, while only the Japanese versions were used in the body of the text. In the resource sections, organization names are presented in Japanese first, followed by the original English in parentheses. In the main body text, however, they are only in Japanese. When the source text includes abbreviations or acronyms that may be unfamiliar to Japanese readers, the full form is translated into Japanese. Physical and email addresses, URLs, and phone numbers are left in their original English form.

For typographic conventions, half-width characters are used for Arabic numerals, the percentage sign, slashes for fractions and acronyms, and colons (where unavoidable). Full-width characters are used for exclamation marks, question marks, Japanese middle dots, slashes (except in the cases mentioned above), ampersands, and parentheses.
The UTF-8 encoding used in our experiments preserves all the relevant features of Japanese grammar and notation.

\section{Runtime Details}\label{section:RuntimeDetails}

\subsection{MT and LLM translation runtimes and costs}

The available runtime and cost details for our translation operations are provided in Table \ref{tab:mt_llm_runtimes} below.

\subsection{“Bug-free” vs. “buggy” prompts}

As noted in Section \ref{subsection:buggy}, we generated two sets of LLM outputs for \texttt{5\_en\_short} (229 segments) with the “bug-free” prompt to compare them with the outputs for the “buggy” prompt and computed their automatic scores with MATEO. See Tables \ref{tab:combined_outputs_russian} and \ref{tab:combined_outputs_japanese} below.

\section{Automatic Quality Evaluation Details}\label{section:auto-quality-details}

As noted in Section \ref{subsection:goals_first_article}, evaluating the quality of MT output is a central concern in research and development. Automatic MT quality metrics are tools that help measure how good an MT-translated sentence is, typically by comparing it to one or more human reference translations. While the field is rapidly evolving, several widely used metrics include BLEU, chrF, TER, and COMET. These can be broadly categorized into \emph{string-based} and \emph{neural-based} metrics.

String-based metrics evaluate translations by comparing the surface forms---words or characters---of MT output and reference human translations. BLEU (BiLingual Evaluation Understudy, \citealp{papineni-etal-2002-bleu}) is one of the earliest and most well-known metrics. BLEU calculates how many $n$-grams (word sequences) in the MT output match those in the reference. While useful, it can be overly strict, penalizing valid translations that use synonyms or different phrasing. chrF (Character F-score, \citealp{popovic-2015-chrf}), on the other hand, operates at the character level, making it more sensitive to morphologically rich languages and spelling. It computes $F$-scores based on overlapping character $n$-grams, which helps in capturing partial matches more effectively. TER (Translation Edit Rate, \citealp{snover-etal-2006-study}) metric measures the number of edits (insertions, deletions, substitutions, and shifts) needed to change the MT output into the reference translation. A lower TER indicates better translation quality. It gives a more intuitive sense of the editing effort required.

Neural-based metrics leverage LLMs and machine learning techniques to evaluate translations more like humans do. These models can understand meaning beyond surface similarity. One such metric, used in this paper, is COMET (Crosslingual Optimized Metric for Evaluation of Translation, \citealp{rei-etal-2020-comet}). Built on pre-trained neural models and fine-tuned on human quality assessments, it can capture semantic similarity and fluency better than traditional metrics, even when there is little word overlap. It has been shown to correlate better with human judgments. Our results reported in Section \ref{subsection:Correlation} are consistent with this claim.

Machine translation evaluation is a fast-moving area of research, with new methods and tools emerging regularly. A great place to stay updated is \url{http://www.machinetranslate.org}, which offers accessible summaries of research, tools, and best practices in the field.

As noted in Section \ref{subsection:goals_first_article}, one exciting development for practitioners is that freelance translators and non-specialists can now use web-based tools to evaluate MT output themselves. \mbox{MATEO} \citep{vanroy-etal-2023-mateo}, employed in our work, is a user-friendly \href{https://streamlit.io/}{Streamlit}-based \href{https://mateo.ivdnt.org/Evaluate}{platform} that allows anyone to calculate multiple MT quality metrics—including BLEU, chrF, TER, and COMET—without needing technical knowledge.

There is thus no mystery to MT quality evaluation. These metrics, whether simple or complex, are just tools to help us understand how well an NMT engine or an LLM has translated a piece of text. As the tools become more accessible and sophisticated, translators and content creators are empowered to make informed decisions about using and improving MT output.

A point of caution: when utilizing MATEO, or any other toolkit, for MT evaluation, it is crucial to select the appropriate \emph{metric configurations}{} to ensure accurate and meaningful results. While the default settings in MATEO are designed to be practical for a wide range of target languages, evaluating translations into morphologically rich languages, such as Japanese or Korean, requires special attention. These languages exhibit complex word forms and inflections that standard metric configurations might not fully capture.

MATEO allows one to make the necessary changes in the “Metric selection” section of the application. In our case, evaluation of the outputs in Japanese required changing the default tokenization setting in BLEU to \texttt{ja-mecab}, and enabling \texttt{asian-support} in TER. We also found it useful to enable the \texttt{normalized} mode in TER, which is set to ‘False' by default.

We report the configurations we used for the automatic evaluation metrics for both our target languages in Tables \ref{tab:en_ru_auto_metrics_config} and \ref{tab:en_ja_auto_metrics_config}.

\section{Evaluation Scores for Sub-Documents of Different Sizes}\label{section:eval-scores-diff-sizes}

See Tables \ref{tab:evaluation_scores} and \ref{tab:evaluation_scores_japanese} below, which represent the four sets of automatic metric scores for three non-overlapping parts of \texttt{1-8\ en short} along with the whole: 3555 = 228 + 1143 + 2183 segments.

\section{Pearson Correlations for Three Pairs of Score Value Sets Across Translation Systems}\label{section:pearson-three}

Pearson’s correlation coefficient ($r$) is a statistical measure that quantifies the strength and direction of the linear relationship between two variables, helping to determine whether changes in one variable are associated with changes in another.

The formula for calculating Pearson’s correlation is:

$$r = \frac{\sum\limits_{i=1}^{n} (x_i - \bar{x})(y_i - \bar{y})}{\sqrt{\sum\limits_{i=1}^{n} (x_i - \bar{x})^2 \sum\limits_{i=1}^{n} (y_i - \bar{y})^2}}$$

\noindent where $x_i$ and $y_i$ are individual data points, $\bar{x}$ and $\bar{y}$ are their means, and $n$ is the number of data pairs. In essence, $r$ measures how much two variables change together relative to how much they change individually. The numerator represents the covariance between the variables, while the denominator normalizes this value using the standard deviations of both variables. 

The value of $r$ always lies between $-1.0$ (corresponding to perfect negative correlation) and 1.0 (perfect positive correlation). Typical guidelines for interpreting the values of $r$ are as follows:

\begin{table}[H]
\centering
\renewcommand{\arraystretch}{1.2} 
\setlength{\tabcolsep}{6pt} 
\begin{tabular}{|c|c|}
\hline
\textit {\textbf{r}} & \textbf{Correlation} \\ \hline
0.9 to 1.0 or $-0.9$ to $-1.0$ & Very strong \\ \hline
0.7 to 0.9 or $-0.7$ to $-0.9$ & Strong \\ \hline
0.5 to 0.7 or $-0.5$ to $-0.7$ & Moderate \\ \hline
0.3 to 0.5 or $-0.3$ to $-0.5$ & Weak \\ \hline
0.0 to 0.3 or 0.0 to $-0.3$ & Very week or none \\ \hline

\end{tabular}
\end{table}

The $p$-value associated with Pearson’s correlation $r$ estimates the statistical significance of the observed correlation. A $p$-value is the probability that the actual distribution of the data points would occur by random chance. A low $p$-value (typically $< 0.05$) suggests that the result is statistically significant.

In our case, the variables in questions are pairwise metric-specific scores which are reflected in the rows of Tables \ref{tab:evaluation_scores} and \ref{tab:evaluation_scores_japanese}. For example, the first two rows in Table \ref{tab:evaluation_scores} (for EN-RU) show the COMET scores for \texttt{229\_en} and \texttt{1143\_en} across our six MT/LLM systems. Accordingly: $x_1 = 87.7, x_2 = 91.1, x_3 = 90.0, x_4 = 88.7, x_5 = 89.6, x_6 = 88.9, y_1 = 89.4, y_2 = 91.4, y_3 = 90.8, y_4 = 89.7, y_5 = 90.1, y_6 = 89.1,$  yielding $r = 0.891; p = 0.0077$, reflected in the last column of Table \ref{tab:correlations}.

Thus Tables \ref{tab:correlations} and \ref{tab:correlations_enja} below display the correlation coefficients and their $p$-values for three pairs of score value sets for the outputs from our range of six translation systems (i.e. MT1–MT3 and LLM1–LLM3), in both language directions.

\begin{figure*}[htbp]
\centering
\includegraphics[width=\linewidth]{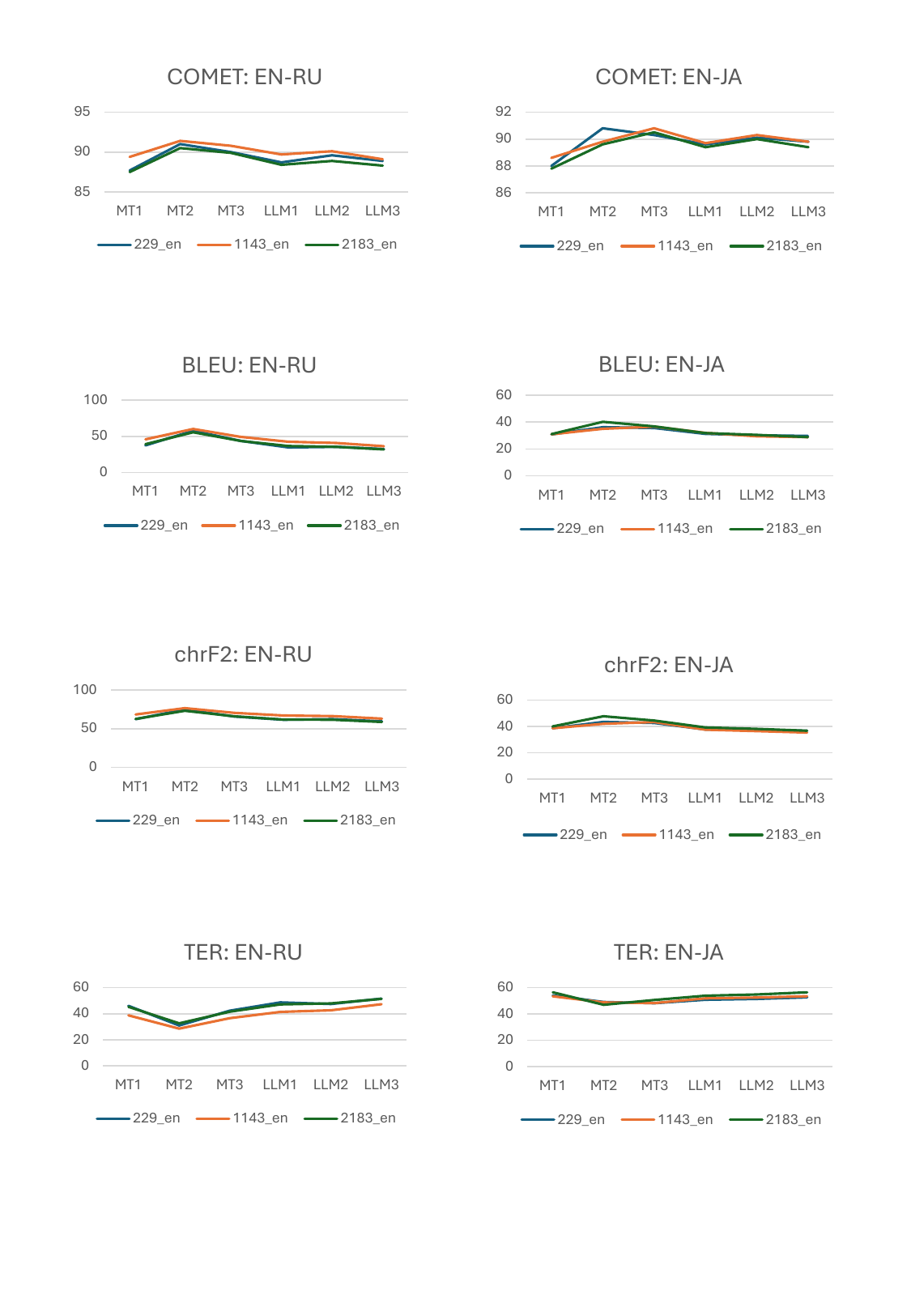}
\caption{Automatic evaluation scores across the outputs of six translation systems for three non-overlapping parts of the RFTC corpus: 229, 1143, and 2183 segments.}
\label{fig:diff-size-plots}
\end{figure*}

The plots in Figure \ref{fig:diff-size-plots} provide the additional details of the distribution of our “data points” across MT1–3 and LLM 1–3.

As we noted in Section \ref{section:SampleSize}, all the pairwise Pearson correlations for our three non-overlapping sub-documents are very strong and statistically significant thus highlighting the consistency and stability of the rankings of our MT/LLM outputs across sub-documents of different sizes. If we wanted to select one or two best performing systems based on the automatic evaluation scores for our project, we could simply pick out the shortest chapter of PRG (i.e. Chapter 5 = \texttt{229\_en}) and treat it as a good representative of the entire document.

Even this shortest sample has over 4,000 source words, which exceeds the average daily output of a typical translator. It would be interesting to trim down the sample size even more to determine the point at which the correlation is lost and the scores become unreliable. The best way to do this is to perform a \emph{power analysis} using one of the available toolkits (e.g. \citealp{zhu-etal-2020-nlpstattest}). It would also be desirable to include contrasting pairs of data points from different translation domains and registers. We leave it for further work.

Translators interested in implementing correlation or more nuanced statistical analyses can use any number of generally available tools, from Excel to Python or R libraries. In our experience, LLMs can generate simple standalone Python scripts for such purposes, in response to sufficiently detailed prompts.

\section{Manual Grading Scale}\label{section:manual-grading-scale}

Our manual scale modeled after academic grading is displayed in Table \ref{tab:grading_scale}.

\section{Correlation Between Sentence-Level COMET Scores and Numeric Human Grades}\label{section:human-comet-correlation}

Tables \ref{tab:pearson_spearman} and \ref{tab:pearson_spearman2} represent Pearson and Spearman correlation between sentence-level COMET scores and our numeric human grades.

As noted above (Appendix \ref{section:pearson-three}), Pearson correlation measures the strength and direction of a \emph{linear} relationship between two variables. It assumes that both variables are normally distributed, and that the relationship is linear. This approximation was adequate for six pairs of data points representing the scores for the outputs of our MT/LLM systems. But the number of our chosen sentence-level COMET scores and the corresponding human grades is larger: 30 or 60. In such cases Pearson correlation may be insufficient, especially if the relationship between variables is non-linear or if the data contains outliers. In such cases, adding Spearman correlation ($\rho$) can provide a more accurate picture of the association by focusing on the rank-order rather than precise values. Spearman correlation is a non-parametric measure that assesses how well the relationship between two variables can be described by a \emph{monotonic} function. It uses the ranked values of the data, not the raw values, so it doesn’t assume normality or linearity.

\section{APIs and Applied Technical Avenues for Freelancers}\label{section:apis-applied-technical-avenues-freelancers}

We acknowledge that most freelance translators are not programmers. However, as discussed elsewhere in this paper, we believe the future of translation work demands skills that are more conducive to the building blocks of machine intelligence and automation. 

As a practical matter, the major LLM providers expose use of their models in two ways: a human-friendly way, via an interactive “chatbot” interface, and a machine-friendly way, via REST (REpresentational State Transfer) APIs, or Application Programming Interfaces. Despite the imposing weight of these acronyms to non-technical readers, the chasm between these modes of interaction is not, in fact, so vast. REST APIs use HTTP, the building-block protocol of the World Wide Web, as a transport, and a series of HTTP chatbot “verbs” whose meaning is not especially obscure: \texttt{GET}, \texttt{POST}, \texttt{DELETE}, and so on. 

Contemporary REST APIs customarily encode information in a lightweight, human-readable encapsulation structure known as \href{https://www.json.org}{JSON} (JavaScript Object Notation). The primary purpose of JSON is to define a hierarchical relational structure---for instance, to distinguish an object from its attributes. 

A simple JSON structure might look like this:

\begin{lstlisting}
    {
        "people": [
            "Alex": {
                "org": "Evariste Systems",
                "phd": false
            },
            "Yuri": {
                "org": "University of Georgia",
                "phd": true
            }
        ],
        "paper_type": "edifying",
        "lucky_numbers": [1, 7, 10]
    }
\end{lstlisting}

The SDKs (Software Development Kits) of major LLM providers abstract away lower-level programmatic REST API interactions, which is more ergonomic for the software engineers using them. However, the LLM APIs can be directly queried, with the help of user-friendly tools such as \href{https://www.postman.com}{Postman}. The interactive “chatbot” clients to which many readers will be well-accustomed are little more than simplified front-ends to these REST APIs.

Perusing the content of JSON responses from the major LLM providers' REST APIs can open one's mind to new possibilities. For example, one of the authors used Postman to prompt OpenAI's GPT-4o model thus:

“You are a highly competent Russian to English translator. How would you explain the Russian concept of a ‘matryoshka' in English? Please be brief.”

On the surface, the reply received was unremarkable:

\begin{lstlisting}
    {
  "id": "chatcmpl-BLGsJZWEEWEfIBCkS46cXpNyyTfnY",
  "object": "chat.completion",
  "created": 1744409455,
  "model": "gpt-4o-2024-08-06",
  "choices": [
    {
      "index": 0,
      "message": {
        "role": "assistant",
        "content": "A matryoshka, also known as a Russian nesting doll, is a set of wooden dolls of decreasing size placed one inside another. Each doll splits in half at the middle to reveal a smaller doll inside, symbolizing themes of motherhood, family, and continuity.",
        "refusal": null,
        "annotations": []
      },
      [...]
\end{lstlisting}

However, after perusing \href{https://platform.openai.com/docs/api-reference/chat/create}{OpenAI chat API reference}, the author learned that it is possible to supply the JSON attributes:

\begin{lstlisting}
    "logprobs": true,
    "top_logprobs: 3
\end{lstlisting}

\noindent to the request, which tells OpenAI to share two other alternative probabilistic paths not taken for every generated token. 

Thus, although GPT-4o began this generated response with the article ‘A’, it considered alternatives:

\begin{lstlisting}
    "logprobs": {
        "content": [
          {
            "token": "A",
            "logprob": -0.011159946210682392,
            "bytes": [
              65
            ],
            "top_logprobs": [
              {
                "token": "A",
                "logprob": -0.011159946210682392,
                "bytes": [
                  65
                ]
              },
              {
                "token": "The",
                "logprob": -4.511159896850586,
                "bytes": [
                  84,
                  104,
                  101
                ]
              },
              {
                "token": "In",
                "logprob": -9.636159896850586,
                "bytes": [
                  73,
                  110
                ]
              }
            ]
          },
\end{lstlisting}

The author fed this probability output for the first few tokens into Anthropic’s Claude Sonnet model and asked it to generate a flowchart, using the following prompt: 

\begin{quote}
“This JSON file contains a ‘logprobs’ element from OpenAI's API, which shows two other probabilistic responses considered by the model before returning the one with the lowest absolute value. Could you please draw a flowchart for the first five ‘logprobs’ elements which illustrates the traversal path taken? Please encapsulate every one of the five generated tokens in a rectangle, and use a darker or solid line to indicate the path actually taken based on the lowest absolute value of the ‘logprob’ entry, while using lighter lines, their lightness in proportion to the relative absolute value of the ‘logprob’ value, to show alternatives not taken.”
\end{quote}

The author then further prompted Claude to flatten the graphic for ease of inclusion here:

\begin{quote}
    “Could you refine this flowchart to be more vertical, so that it is easier to incorporate into a two-column document without overflow beyond the margins?”
\end{quote}

This was the result:

\begin{center}
    \includegraphics[width=0.9\columnwidth]{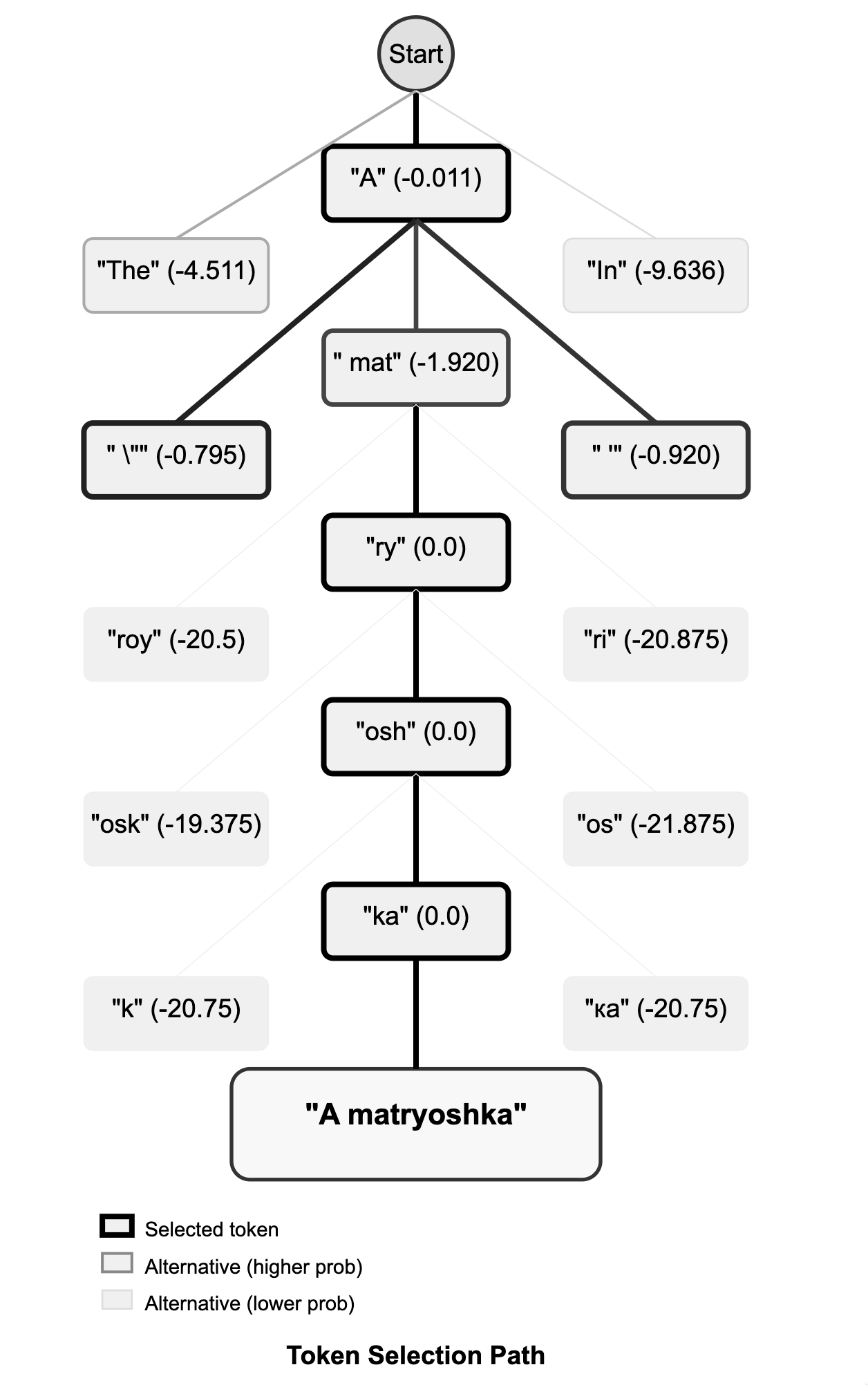}
\end{center}

A key idea here is that this interesting foray would have never occurred to the author without digging into the OpenAI APIs and interacting with them directly. The commonplace interactive chatbot interfaces do not surface these possibilities to the end-user. No code was written for this exercise, just some slight tweaks to minimal, easy-to-read JSON data structures.

First and foremost, we believe that becoming conversant with the API surface of the major LLM providers can empower freelancers to make more specific and technically articulate LLM integration demands of the vendors of their preferred translation software tools. Second, we can reasonably speculate that the creation or enhancement of tools reliant on API integrations may drift further away from the exclusive province of professional programmers, and become more reachable for technically minded end-users. This trend can be extrapolated from an ongoing trajectory to which veteran software engineers are privy: service APIs offered over the Internet have become far less arcane and easier to decipher over time, between the simplified vocabulary of REST and the human-readable wire format of JSON, for example.

\section{Carbon Imprint}\label{section:carbon-imprint}

We used the Green Algorithm developed in \citep{lannelongue_green_2021}  to estimate the carbon imprint of our computations performed on two computers (Figure \ref{fig:carbon1} and Figure \ref{fig:carbon2} below).

\FloatBarrier

\begin{table*}[htbp]
\centering
\renewcommand{\arraystretch}{1.3} 
\setlength{\tabcolsep}{8pt} 
\begin{tabular}{|l|c|c|c|}
\hline
\textbf{}                          & \textbf{\texttt{1-10\_en\_long}} & \textbf{\texttt{1-10\_en\_short}} & \textbf{\texttt{1-8\_en\_short}} \\ \hline
Segments                  & \textbf{4528}                                        & \textbf{3896}                                        & \textbf{3555}                                       \\ \hline
Words tokens (no punc\textbf{)}    & 76,553                                       & 74,667                                       & 68,989                                      \\ \hline
Word types (no punc)      & 10,689                                       & 10,325                                       & 9,821                                       \\ \hline
Characters (w/o \textbackslash r\textbackslash n) & 500,347                                    & 485,830                                     & 448,652                                     \\ \hline
Type/token ratio          & 0.14                                         & 0.14                                        & 0.14                                        \\ \hline

MTLD          & 100.18                                         & 101.44                                        & 100.77                                        \\ \hline

Average segment length (words) & 16.91                                    & 19.17                                       & 19.41                                       \\ \hline
Average word length (characters) & 5.42                                    & 5.38                                        & 5.38                                        \\ \hline
\end{tabular}
\caption{Source document statistics.}
\label{tab:source_document_statistics}
\end{table*}

\FloatBarrier

\FloatBarrier 

\begin{table*}[ht]
\centering
\begin{tabular}{@{}>{\RaggedRight\arraybackslash}p{4.5cm}rrrrr@{}}
\toprule
\textbf{Type}                             & \textbf{Segments.} & \textbf{Source words} & \textbf{Source chars} & \textbf{Source tags} & \textbf{Percent} \\ \midrule
\multicolumn{6}{c}{\textbf{\texttt{1-10\_en\_short}}} \\ \midrule
All                                       & \textbf{3896}      & 74683                 & 415016                & 0                    & 100              \\
X-translated / double context             & 0                  & 0                     & 0                     & 0                    & 0                \\
Repetition                                & 0                  & 0                     & 0                     & 0                    & 0                \\
101\%                                     & 0                  & 0                     & 0                     & 0                    & 0                \\
100\%                                     & 0                  & 0                     & 0                     & 0                    & 0                \\
95\%--99\%                                & 3                  & 31                    & 227                   & 0                    & 0.04             \\
85\%--94\%                                & 6                  & 118                   & 696                   & 0                    & 0.16             \\
75\%--84\%                                & 17                 & 251                   & 1703                  & 0                    & 0.34             \\
50\%--74\%                                & 201                & 3080                  & 18615                 & 0                    & 4.12             \\
No match                                  & 3669               & 71203                 & 393775                & 0                    & 95.34            \\ \midrule
\multicolumn{6}{c}{\textbf{\texttt{1-8\_en\_short}}} \\ \midrule
All                                       & \textbf{3555}      & 68999                 & 383178                & 0                    & 100              \\
X-translated / double context             & 0                  & 0                     & 0                     & 0                    & 0                \\
Repetition                                & 0                  & 0                     & 0                     & 0                    & 0                \\
101\%                                     & 0                  & 0                     & 0                     & 0                    & 0                \\
100\%                                     & 0                  & 0                     & 0                     & 0                    & 0                \\
95\%--99\%                                & 3                  & 31                    & 227                   & 0                    & 0.04             \\
85\%--94\%                                & 6                  & 118                   & 696                   & 0                    & 0.17             \\
75\%--84\%                                & 16                 & 239                   & 1636                  & 0                    & 0.35             \\
50\%--74\%                                & 172                & 2700                  & 16465                 & 0                    & 3.91             \\
No match                                  & 3358               & 65911                 & 364154                & 0                    & 95.52            \\ \bottomrule
\end{tabular}

\caption{\label{tab:memoQanalysis}
MemoQ analysis of the “long” and “short” PRG inputs. Fuzzy matches result from the \textit{Homogeneity} feature of the memoQ analysis which measures internal similarities within a set of documents by adding each segment to a temporary TM and using it for lookup for every subsequently processed segment. For details, see \href{https://docs.memoq.com/current/en/Concepts/concepts-homogen-rep-project.html}{here}.}

\end{table*}

\FloatBarrier

\FloatBarrier 

\begin{table*}[!htbp] 
\centering
\renewcommand{\arraystretch}{1.2} 
\setlength{\tabcolsep}{4pt} 
\begin{tabular}{|p{6cm}|p{6cm}|p{4cm}|}
\hline
\textbf{Document Name}                      & \textbf{Excel/PDF}                                       & \textbf{Notes}                                   \\ \hline
\textbf{PARALYSIS RESOURCE GUIDE (PRG): International Edition}    & 1-10\_en-ru-ja\_long.xlsx                                    & 4528 segments                                    \\
                                            &                                     & (ID: 3--4687)                                   \\
                                            & 1-10\_en-ru-ja\_short.xlsx                                   & 3896 segments                                   \\
                                            &                                     & (ID: 13--4687)                                  \\
                                            & PRG-IntEd\_en.pdf                                             &                                                 \\
                                            & PRG-IntEd\_ru.pdf                                             &                                                 \\
                                            & PRG-IntEd\_ja.pdf                                             &                                                 \\ \hline
\textbf{PRG: Chapters 1-6}                 & 1-8\_en-ru-ja\_short.xlsx                                    & 3555 segments                                   \\
                                            &                                                             & (ID: 13--4299)                                  \\ \hline
\textbf{PRG: Front Cover}                  &                                                             & No segments                                     \\ \hline
\textbf{PRG: Introduction}                 & 2\_en-ru-ja\_short.xlsx                                      & 21 segments                                     \\
                                            &                                                             & (ID: 13--45)                                    \\ \hline
\textbf{PRG: Chapter 1}                    & 3\_en-ru-ja\_short.xlsx                                      & 1143 segments                                   \\
                                            &                                        & (ID: 108--1430)                                 \\ \hline
\textbf{PRG: Chapter 2}                    & 4\_en-ru-ja\_short.xlsx                                      & 1237 segments                                   \\
                                            &                                                             & (ID: 1437--2897)                                \\ \hline
\textbf{PRG: Chapter 3}                    & 5\_en-ru-ja\_short.xlsx                                      & 229 segments                                    \\
                                            &                                        & (ID: 2903--3178)                                \\ \hline
\textbf{PRG: Chapter 4}                    & 6\_en-ru-ja\_short.xlsx                                      & 228 segments                                    \\
                                            &                                                             & (ID: 3186--3467)                                \\ \hline
\textbf{PRG: Chapter 5}                    & 7\_en-ru-ja\_short.xlsx                                      & 583 segments                                    \\
                                            &                                                             & (ID: 3473--4156)                                \\ \hline
\textbf{PRG: Chapter 6}                    & 8\_en-ru-ja\_short.xlsx                                      & 114 segments                                    \\
                                            &                                                             & (ID: 4164--4299)                                \\ \hline
\textbf{PRG: Glossary}                     & 9\_en-ru-ja\_short.xlsx                                      & 335 segments                                    \\
                                            &                                                             & (ID: 4306--4673)                                \\ \hline
\textbf{PRG: Back Cover}                   & 10\_en-ru-ja\_short.xlsx                                     & 6 segments                                      \\
                                            &                                                             & (ID: 4675--4687)                                \\ \hline
\end{tabular}
\caption{\label{tab:prg_table}
Document details for the Paralysis Resource Guide (PRG).}
\end{table*}

\FloatBarrier 

\FloatBarrier
\begin{table*}[htbp]
\centering
\renewcommand{\arraystretch}{1.2} 
\setlength{\tabcolsep}{8pt} 
\begin{tabular}{|l|l|c|c|l|}
\hline
\textbf{}          & \textbf{}          & \textbf{Runtime} & \textbf{Total cost} & \textbf{Notes}                               \\ \hline
\textbf{MT1}       & EN-RU              & 00:02:17         & n/a                 &                                              \\ 
                   & EN-JA              & 00:02:32         & n/a                 &                                              \\ \hline
\textbf{MT2}       & EN-RU              & 00:01:28         & n/a                 &                                              \\ 
                   & EN-JA              & 00:01:05         & n/a                 &                                              \\ \hline
\textbf{MT3}       & EN-RU              & 00:08:36         & n/a                 &                                              \\ 
                   & EN-JA              & 00:08:03         & n/a                 &                                              \\ \hline
\textbf{LLM1}      & EN-RU              & 00:15:38    & USD 7.01           & Combined EN-RU and EN-JA                    \\ 
                   & EN-JA              & 00:15:28    &                    &                                              \\ \hline
\textbf{LLM2}      & EN-RU              & 02:29:43    & USD 23.22          & Combined EN-RU and EN-JA                    \\ 
                   & EN-JA              & 02:38:38    &                    &                                              \\ \hline
\textbf{LLM3}      & EN-RU              & 00:11:25    & n/a                &                                              \\ 
                   & EN-JA              & 00:11:25    &                    &                                              \\ \hline
\end{tabular}
\caption{MT and LLM translation runtimes and costs for \texttt{1-10\_en\_short} (3896 segments).}
\label{tab:mt_llm_runtimes}
\end{table*}

\begin{table*}[h]
\centering
\small
\renewcommand{\arraystretch}{1.3} 
\setlength{\tabcolsep}{1pt} 
\begin{tabular}{|l|c|c|c|c|}
\hline
\textbf{}            & \textbf{COMET}          & \textbf{BLEU}           & \textbf{chrF2}          & \textbf{TER}            \\ \hline
\textbf{“Bug Free-1”} & $89.1 \pm 0.4$         & $33.6 \pm 1.7$          & $60.7 \pm 1.2$          & $49.6 \pm 1.5$          \\ \hline
\textbf{“Bug Free-2”} & $89.0 \pm 0.5$         & $33.7 \pm 1.7$ ($p=0.23$)* & $60.7 \pm 1.1$ ($p=0.33$)* & $49.6 \pm 1.5$ ($p=0.33$) \\ \hline
\textbf{“Buggy”}      & $89.1 \pm 0.5$ ($p=0.42$) & $34.1 \pm 1.7$ ($p=0.08$) & $61.0 \pm 1.2$ ($p=0.04$)* & $49.3 \pm 1.5$ ($p=0.09$) \\ \hline
\end{tabular}
\caption{LLM1--3 combined outputs for \texttt{5\_en\_short}: English-Russian.}
\label{tab:combined_outputs_russian}
\end{table*}

\begin{table*}[h]
\centering
\small
\renewcommand{\arraystretch}{1.3} 
\setlength{\tabcolsep}{1pt} 
\begin{tabular}{|l|c|c|c|c|}
\hline
\textbf{}            & \textbf{COMET}          & \textbf{BLEU}           & \textbf{chrF2}          & \textbf{TER}            \\ \hline
\textbf{“Bug Free-1”} & $89.7 \pm 0.4$         & $29.8 \pm 1.3$          & $36.8 \pm 1.3$          & $52.0 \pm 1.3$          \\ \hline
\textbf{“Bug Free-2”} & $89.7 \pm 0.4$ ($p=0.34$) & $29.6 \pm 1.3$ ($p=0.15$) & $36.4 \pm 1.2$ ($p=0.05$)* & $52.2 \pm 1.3$ ($p=0.09$) \\ \hline
\textbf{“Buggy”}      & $89.9 \pm 0.4$ ($p=0.05$)* & $30.5 \pm 1.3$ ($p=0.02$)* & $37.4 \pm 1.3$ ($p=0.04$)* & $51.6 \pm 1.3$ ($p=0.05$) \\ \hline
\end{tabular}
\caption{LLM1--3 combined outputs for \texttt{5\_en\_short}: English-Japanese.}
\label{tab:combined_outputs_japanese}
\end{table*}

\begin{table*}[h]
\centering
\scriptsize
\renewcommand{\arraystretch}{1.3} 
\setlength{\tabcolsep}{5pt} 
\begin{tabular}{|l|l|}
\hline
\textbf{Metric} & \textbf{Details} \\ \hline
BLEU   & \texttt{nrefs:1|bs:1000|seed:12345|case:mixed|eff:no|tok:13a|smooth:exp|version:2.3.1|mateo:1.1.3} \\ \hline
chrF2  & \texttt{nrefs:1|bs:1000|seed:12345|case:mixed|eff:yes|nc:6|nw:0|space:no|version:2.3.1|mateo:1.1.3} \\ \hline
TER    & \texttt  {nrefs:1|bs:1000|seed:12345|case:lc|tok:tercom|norm:yes|punct:yes|asian:no|version:2.3.1|mateo:1.1.3} \\ \hline
COMET  & \texttt{nrefs:1|bs:1000|seed:12345|c:Unbabel/wmt22-comet-da|version:2.0.1|mateo:1.1.3} \\ \hline
\end{tabular}
\caption{Metrics configurations for English-Russian.}
\label{tab:en_ru_auto_metrics_config}
\end{table*}

\begin{table*}[h]
\centering
\scriptsize
\renewcommand{\arraystretch}{1.3} 
\setlength{\tabcolsep}{5pt} 
\begin{tabular}{|l|l|}
\hline
\textbf{Metric} & \textbf{Details} \\ \hline
BLEU   & \texttt{nrefs:1|bs:1000|seed:12345|case:mixed|eff:no|tok:ja-mecab-0.996-IPA|smooth:exp|version:2.3.1|mateo:1.1.3} \\ \hline
chrF2  & \texttt{nrefs:1|bs:1000|seed:12345|case:mixed|eff:yes|nc:6|nw:0|space:no|version:2.3.1|mateo:1.1.3} \\ \hline
TER    & \texttt{nrefs:1|bs:1000|seed:12345|case:lc|tok:tercom|norm:yes|punct:yes|asian:yes|version:2.3.1|mateo:1.1.3} \\ \hline
COMET  & \texttt{nrefs:1|bs:1000|seed:12345|c:Unbabel/wmt22-comet-da|version:2.0.1|mateo:1.1.3} \\ \hline
\end{tabular}
\caption{Metrics configurations for English-Japanese.}
\label{tab:en_ja_auto_metrics_config}
\end{table*}

\FloatBarrier
\begin{table*}[h]
\centering
\renewcommand{\arraystretch}{1.2} 
\setlength{\tabcolsep}{4pt} 
\begin{tabular}{llccc}
\toprule
       &              & \textbf{BLEU-chrF2} & \textbf{BLEU-TER} & \textbf{BLEU-COMET} \\ \midrule
\textbf{EN-RU} & $r$          & 0.998             & -0.999            & 0.806              \\
               & $p$          & \textless 0.0001          & \textless 0.0001          & 0.0345             \\ \midrule
\textbf{EN-JA} & $r$          & 0.990             & -0.969            & 0.388              \\
               & $p$          & \textless 0.0001          & 0.0002            & 0.4321             \\ 
\bottomrule
\end{tabular}
\caption{Pearson correlations ($r$) for BLEU-chrF2, BLEU-TER, and BLEU-COMET for \texttt{1-8\_en}.}
\label{tab:pearson_correlations}
\end{table*}

\begin{table*}[h]
\centering
\small
\renewcommand{\arraystretch}{1.5}
\setlength{\tabcolsep}{7pt}
\begin{tabular}{|l|c|c|c|c|c|c|c|c|}
\hline
\textbf{Metric} & \textbf{Label} & \textbf{Sgmts} & \textbf{MT1} & \textbf{MT2} & \textbf{MT3} & \textbf{LLM1} & \textbf{LLM2} & \textbf{LLM3} \\ \hline
\textbf{COMET} & \texttt{229\_en}    & 229           & 87.7         & 91.0         & 90.0         & 88.7         & 89.6         & 88.9         \\ 
               & \texttt{1143\_en}    & 1143          & 89.4         & 91.4         & 90.8         & 89.7         & 90.1         & 89.1         \\ 
                & \texttt{2183\_en}    & 2183          & 87.5         & 90.5         & 89.9         & 88.4         & 88.9         & 88.3         \\ 
               & \texttt{1-8\_en}  & 3555          & 88.1         & 90.8         & 90.2         & 88.8         & 89.3         & 88.6         \\ \hline
\textbf{BLEU}  & \texttt{229\_en}    & 229           & 37.2         & 57.7         & 43.6         & 34.6         & 35.3         & 32.3         \\ 
               & \texttt{1143\_en}    & 1143          & 45.8         & 60.1         & 49.0         & 42.5         & 40.8         & 36.2         \\ 
                & \texttt{2183\_en}    & 2183          & 39.0         & 55.5         & 43.6         & 36.5         & 35.3         & 31.8         \\ 
               & \texttt{1-8\_en}  & 3555          & 41.1         & 57.2         & 45.4         & 38.4         & 37.1         & 33.2         \\ \hline
\textbf{chrF2} & \texttt{229\_en}    & 229           & 62.2         & 74.7         & 65.6         & 61.5         & 62.3         & 59.4         \\ 
               & \texttt{1143\_en}    & 1143          & 68.3         & 76.4         & 70.4         & 67.0         & 66.2         & 62.9         \\ 
               & \texttt{2183\_en}    & 2183          & 62.5         & 73.0         & 66.0         & 61.7         & 61.4         & 58.6         \\ 
               & \texttt{1-8\_en}  & 3555          & 64.4         & 74.2         & 67.4         & 63.5         & 63.0         & 60.1         \\ \hline
\textbf{TER}   & \texttt{229\_en}    & 229           & 46.0         & 30.9         & 42.3         & 48.7         & 47.4         & 51.6         \\ 
               & \texttt{1143\_en}    & 1143          & 38.7         & 28.5         & 36.5         & 41.4         & 42.7         & 47.3         \\ 
                & \texttt{2183\_en}    & 2183          & 45.2         & 32.6         & 41.6         & 47.2         & 47.9         & 51.4         \\ 
               & \texttt{1-8\_en}  & 3555          & 43.1         & 31.1         & 40.0         & 45.4         & 46.2         & 50.1         \\ \hline
\end{tabular}
\caption{Evaluation scores for documents of different sizes: English-Russian.}
\label{tab:evaluation_scores}
\end{table*}

\begin{table*}[h]
\centering
\small
\renewcommand{\arraystretch}{1.5}
\setlength{\tabcolsep}{7pt} 
\begin{tabular}{|l|c|c|c|c|c|c|c|c|}
\hline
\textbf{Metric} & \textbf{Label} & \textbf{Sgmts} & \textbf{MT1} & \textbf{MT2} & \textbf{MT3} & \textbf{LLM1} & \textbf{LLM2} & \textbf{LLM3} \\ \hline
\textbf{COMET} & \texttt{229\_en}    & 229           & 88.0         & 90.8         & 90.3         & 89.6         & 90.1         & 89.8         \\ 
               & \texttt{1143\_en}    & 1143          & 88.6         & 89.8         & 90.8         & 89.7         & 90.3         & 89.8         \\ 
 & \texttt{2183\_en}    & 2183          & 87.8         & 89.6         & 90.5         & 89.4         & 90.0         & 89.4         \\                
               & \texttt{1-8\_en}  & 3555          & 88.1         & 89.7         & 90.6         & 89.5         & 90.1         & 89.5         \\ \hline
\textbf{BLEU}  & \texttt{229\_en}    & 229           & 30.8         & 36.3         & 35.7         & 31.3         & 30.3         & 29.8         \\ 
               & \texttt{1143\_en}    & 1143          & 31.0         & 35.1         & 36.7         & 32.0         & 29.5         & 28.9         \\ 
 & \texttt{2183\_en}    & 2183          & 31.2         & 40.3         & 36.9         & 31.9         & 30.6         & 28.9         \\ 
               & \texttt{1-8\_en}  & 3555          & 31.1         & 38.6         & 36.8         & 31.9         & 30.2         & 28.9         \\ \hline
\textbf{chrF2} & \texttt{229\_en}    & 229           & 38.7         & 43.5         & 42.7         & 37.7         & 37.6         & 36.7         \\ 
               & \texttt{1143\_en}    & 1143          & 38.7         & 42.1         & 43.4         & 37.6         & 36.5         & 35.4         \\ 
 & \texttt{2183\_en}    & 2183          & 40.0         & 47.8         & 44.6         & 39.3         & 38.2         & 36.8         \\                
               & \texttt{1-8\_en}  & 3555          & 39.5         & 46.0         & 44.1         & 38.6         & 37.6         & 36.3         \\ \hline
\textbf{TER}   & \texttt{229\_en}    & 229           & 54.1         & 49.2         & 48.2         & 50.9         & 51.4         & 52.7         \\ 
               & \texttt{1143\_en}    & 1143          & 53.4         & 48.8         & 48.3         & 51.9         & 52.6         & 53.5         \\ 

 & \texttt{2183\_en}    & 2183          & 56.5         & 47.0         & 50.7         & 53.9         & 54.9         & 56.5         \\

               & \texttt{1-8\_en}  & 3555          & 55.3         & 47.5         & 49.7         & 53.0         & 53.9         & 55.2         \\ \hline
\end{tabular}
\caption{Evaluation scores for documents of different sizes: English-Japanese.}
\label{tab:evaluation_scores_japanese}
\end{table*}

\begin{table*}[h]
\renewcommand{\arraystretch}{1.3} 
\setlength{\tabcolsep}{4pt} 
\begin{minipage}{0.48\textwidth}
\begin{tabular}{|l|c|c|c|}
\hline
\textbf{EN-RU} & \multicolumn{3}{c|}{\textbf{Correlation pairs}} \\ \hline
               & \textbf{229 / 2183} & \textbf{1143 / 2183} & \textbf{229 / 1143} \\ \hline
\textbf{COMET} &                     &                     &                     \\ \hline
$r$            & 0.978              & 0.942              & 0.891              \\ \hline
$p$-value      & 0.0001             & 0.0014             & 0.0077             \\ \hline
\textbf{BLEU}  & \multicolumn{3}{c|}{}                                    \\ \hline
$r$            & 0.991              & 0.994              & 0.973              \\ \hline
$p$-value      & $<$ 0.0001           & $<$ 0.0001           & 0.0001             \\ \hline
\textbf{chrF2} &                     &                     &                     \\ \hline
$r$            & 0.990              & 0.990              & 0.966              \\ \hline
$p$-value      & $<$ 0.0001           & $<$ 0.0001           & 0.0003             \\ \hline
\textbf{TER}   & \multicolumn{3}{c|}{}                                    \\ \hline
$r$            & 0.992              & 0.990              & 0.971              \\ \hline
$p$-value      & $<$ 0.0001           & $<$ 0.0001           & 0.0002             \\ \hline
\end{tabular}
\caption{Pearson correlations ($r$) and $p$-values for EN-RU for three pairs of score value sets across six translation systems.}
\label{tab:correlations}
\end{minipage}%
\hfill
\begin{minipage}{0.48\textwidth}
\centering
\begin{tabular}{|l|c|c|c|}
\hline
\textbf{EN-JA} & \multicolumn{3}{c|}{\textbf{Correlation pairs}} \\ \hline
               & \textbf{229 / 2183} & \textbf{1143 / 2183} & \textbf{229 / 1143} \\ \hline
\textbf{COMET} &                     &                     &                     \\ \hline
$r$            & 0.869              & 0.989              & 0.797              \\ \hline
$p$-value      & 0.0126             & $<$ 0.0001           & 0.0387             \\ \hline
\textbf{BLEU}  & \multicolumn{3}{c|}{}                                    \\ \hline
$r$            & 0.980              & 0.901              & 0.955              \\ \hline
$p$-value      & 0.0001           & 0.0060             & 0.0007             \\ \hline
\textbf{chrF2} &                     &                     &                     \\ \hline
$r$            & 0.984              & 0.929              & 0.967              \\ \hline
$p$-value      & $<$ 0.0001           & 0.0024             & 0.0003             \\ \hline
\textbf{TER}   & \multicolumn{3}{c|}{}                                    \\ \hline
$r$            & 0.852              & 0.922              & 0.936              \\ \hline
$p$-value      & 0.0174             & 0.0031             & 0.0018             \\ \hline
\end{tabular}
\caption{Pearson correlations ($r$) and $p$-values for EN-JA for three pairs of score value sets across six translation systems.}
\label{tab:correlations_enja}
\end{minipage}
\end{table*}

\begin{table*}[h]
\centering
\renewcommand{\arraystretch}{1.2} 
\setlength{\tabcolsep}{6pt} 
\begin{tabular}{|l|c|}
\hline
\textbf{Letter grade} & \textbf{Numeric grade} \\ \hline
A                    & 4.00                   \\ \hline
A-                   & 3.67                   \\ \hline
A-/B+                & 3.50                   \\ \hline
B+                   & 3.33                   \\ \hline
B                    & 3.00                   \\ \hline
B-                   & 2.67                   \\ \hline
B-/C+                & 2.50                   \\ \hline
C+                   & 2.33                   \\ \hline
C                    & 2.00                   \\ \hline
C-                   & 1.67                   \\ \hline
C-/D+                & 1.50                   \\ \hline
D+                   & 1.33                   \\ \hline
D                    & 1.00                   \\ \hline
F                    & 0.00                   \\ \hline
\end{tabular}
\caption{Manual grading scale.}
\label{tab:grading_scale}
\end{table*}

\begin{table*}[htbp]
\centering
\renewcommand{\arraystretch}{1.5} 
\setlength{\tabcolsep}{8pt} 
\begin{tabular}{@{}lcccccc@{}}
\toprule
\multicolumn{7}{c}{\textbf{\texttt{5\_en\_short}: English-Russian}} \\ \midrule
\textbf{}       & \multicolumn{2}{c}{\textbf{Pearson Correlation}} & \multicolumn{2}{c}{\textbf{Spearman Correlation}} \\
\cmidrule(r){2-3} \cmidrule(l){4-5}
\textbf{}       & $r$  \   & $p$     & $\rho$    & $p$     \\ \midrule
\textbf{MT1}    & 0.689   & $< 0.0001$ & 0.755     & $< 0.0001$ \\
\textbf{MT2}    & 0.549   & 0.0016   & 0.765     & $< 0.0001$ \\
\textbf{MT3}    & 0.660   & 0.0001   & 0.765     & $< 0.0001$ \\
\textbf{LLM1}   & 0.692   & $< 0.0001$ & 0.806     & $< 0.0001$ \\
\textbf{LLM2}   & 0.795   & $< 0.0001$ & 0.734     & $< 0.0001$ \\
\textbf{LLM3}   & 0.548   & 0.0016   & 0.567     & 0.0011   \\ \midrule
\multicolumn{7}{c}{\textbf{\texttt{5\_en\_short}: English-Japanese}} \\ \midrule
\textbf{}       & \multicolumn{2}{c}{\textbf{Pearson Correlation}} & \multicolumn{2}{c}{\textbf{Spearman Correlation}} \\
\cmidrule(r){2-3} \cmidrule(l){4-5}
\textbf{}       & $r$     & $p$     & $\rho$    & $p$     \\ \midrule
\textbf{MT1}    & 0.609   & 0.0004   & 0.743     & $< 0.0001$ \\
\textbf{MT3}    & 0.626   & 0.0002   & 0.699     & $< 0.0001$ \\
\textbf{MT4}    & 0.738   & $< 0.0001$ & 0.657     & 0.0001   \\
\textbf{LLM1}   & 0.827   & $< 0.0001$ & 0.811     & $< 0.0001$ \\
\textbf{LLM2}   & 0.767   & $< 0.0001$ & 0.783     & $< 0.0001$ \\
\textbf{LLM3}   & 0.839   & $< 0.0001$ & 0.910     & $< 0.0001$ \\ \bottomrule
\end{tabular}
\caption{Pearson ($r$) and Spearman ($\rho$) correlation coefficients between sentence-level COMET scores and numeric human grades for the select “10-10-10” translation outputs for \texttt{5\_en\_short}.
\label{tab:pearson_spearman}}
\end{table*}

\begin{table*}[htbp]
\centering
\renewcommand{\arraystretch}{1.5} 
\setlength{\tabcolsep}{8pt} 
\begin{tabular}{@{}lcccccc@{}}
\toprule
\multicolumn{7}{c}{\textbf{\texttt{3\_en\_short}: English-Russian}} \\ \midrule
\textbf{}       & \multicolumn{2}{c}{\textbf{Pearson Correlation}} & \multicolumn{2}{c}{\textbf{Spearman Correlation}} \\
\cmidrule(r){2-3} \cmidrule(l){4-5}
\textbf{}       & $r$         & $p$         & $\rho$      & $p$         \\ \midrule
\textbf{MT1}    & 0.927       & $< 0.0001$  & 0.895       & $< 0.0001$  \\
\textbf{MT2}    & 0.845       & $< 0.0001$  & 0.779       & $< 0.0001$  \\
\textbf{MT3}    & 0.878       & $< 0.0001$  & 0.851       & $< 0.0001$  \\
\textbf{LLM1}   & 0.761       & $< 0.0001$  & 0.830       & $< 0.0001$  \\
\textbf{LLM2}   & 0.697       & $< 0.0001$  & 0.738       & $< 0.0001$  \\
\textbf{LLM3}   & 0.663       & $< 0.0001$  & 0.774       & $< 0.0001$  \\ \midrule
\multicolumn{7}{c}{\textbf{\texttt{3\_en\_short}: English-Japanese}} \\ \midrule
\textbf{}       & \multicolumn{2}{c}{\textbf{Pearson Correlation}} & \multicolumn{2}{c}{\textbf{Spearman Correlation}} \\
\cmidrule(r){2-3} \cmidrule(l){4-5}
\textbf{}       & $r$         & $p$         & $\rho$      & $p$         \\ \midrule
\textbf{MT1}    & 0.558       & $< 0.0001$  & 0.672       & $< 0.0001$  \\
\textbf{MT3}    & 0.831       & $< 0.0001$  & 0.843       & $< 0.0001$  \\
\textbf{MT4}    & 0.630       & $< 0.0001$  & 0.694       & $< 0.0001$  \\
\textbf{LLM1}   & 0.582       & $< 0.0001$  & 0.589       & $< 0.0001$  \\
\textbf{LLM2}   & 0.462       & 0.0002      & 0.436       & 0.0004      \\
\textbf{LLM3}   & 0.665       & $< 0.0001$  & 0.646       & $< 0.0001$  \\ \bottomrule
\end{tabular}
\caption{Pearson ($r$) and Spearman ($\rho$) correlation coefficients between sentence-level COMET scores and numeric human grades for the select “20-20-20” translation outputs for \texttt{3\_en\_short}.}
\label{tab:pearson_spearman2}
\end{table*}

\FloatBarrier
\begin{figure*}[ht]
  \centering
  \begin{minipage}[t]{0.48\textwidth}
    \centering
    \includegraphics[width=\textwidth]{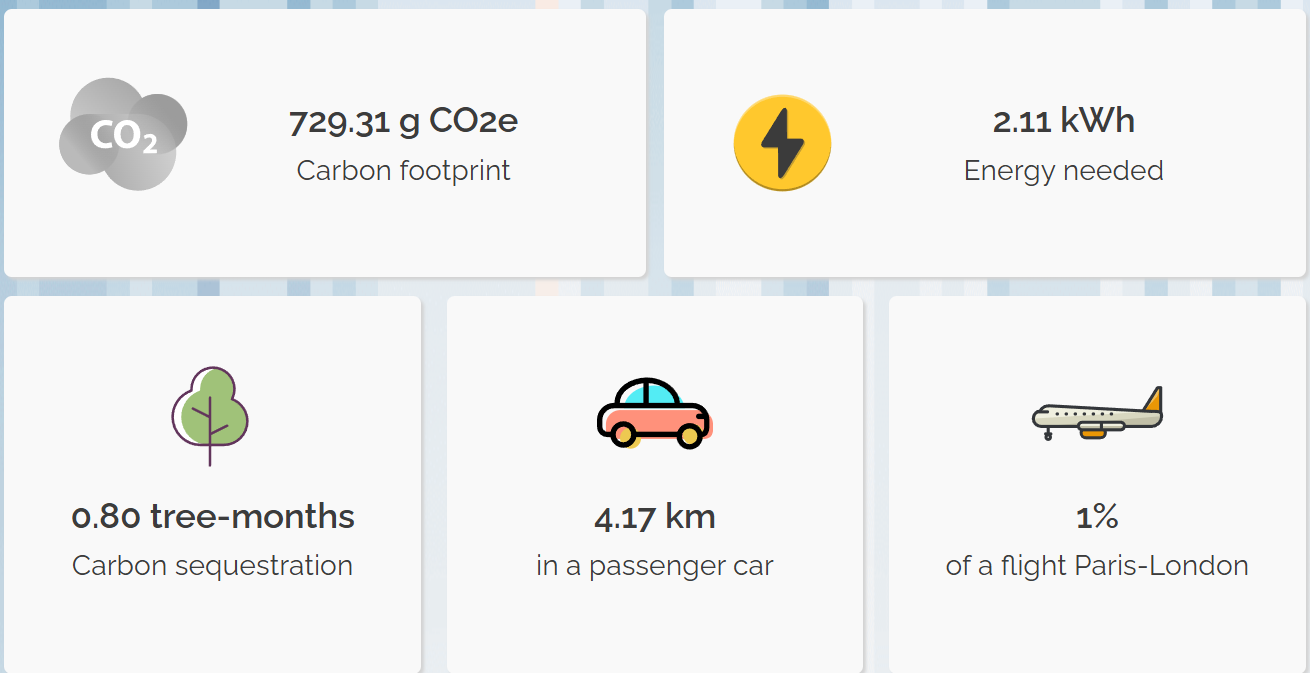}
    \caption{Carbon imprint for 13th Gen Intel(R) Core(TM) i9-13900KF 3.00 GHz 64.0 GB PC.}
    \label{fig:carbon1}
  \end{minipage}
  \hfill
  \begin{minipage}[t]{0.48\textwidth}
    \centering
    \includegraphics[width=\textwidth]{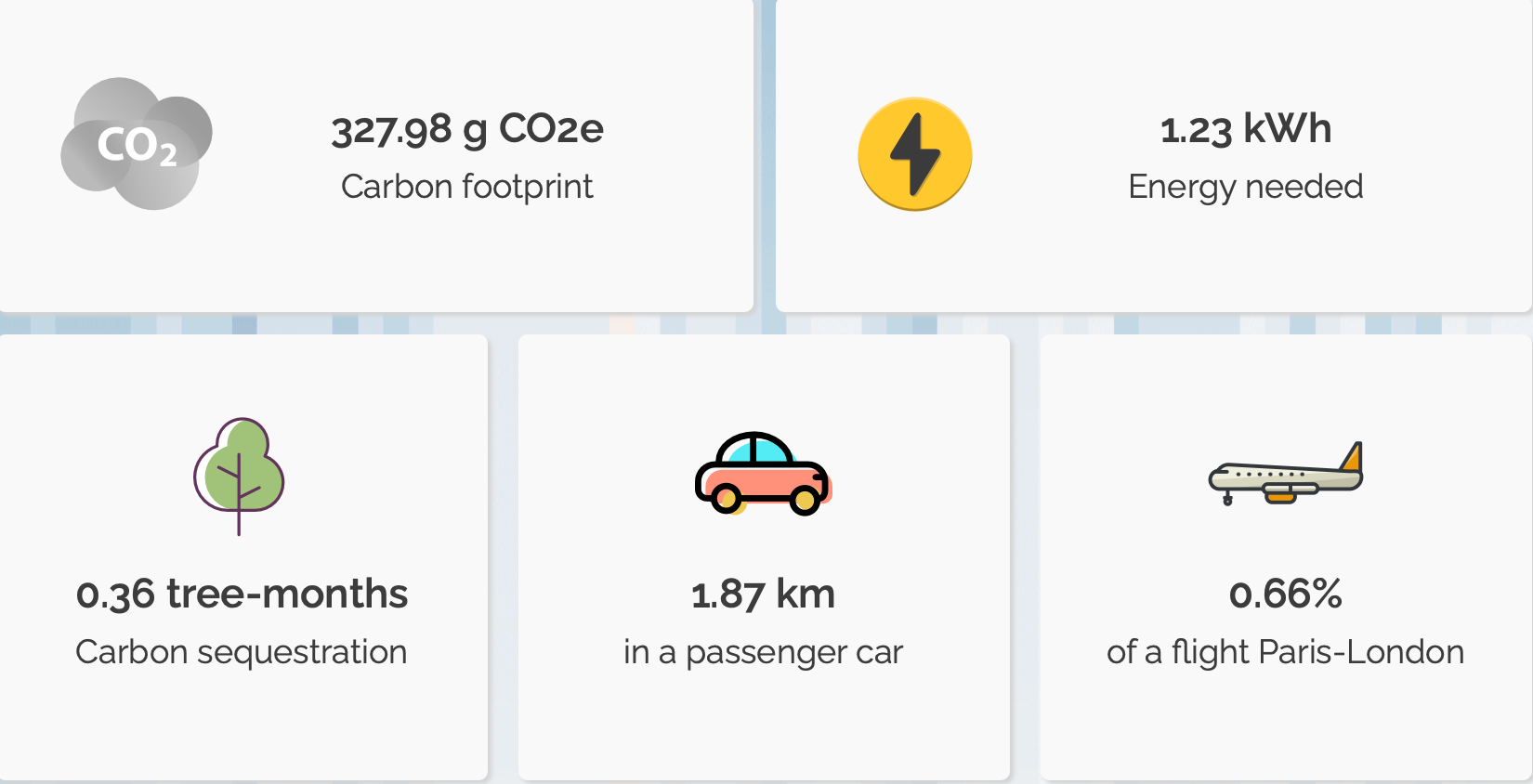}
    \caption{Carbon imprint for virtual server with Intel Xeon (Skylake) 6-core CPU, 16 GB of RAM.}
    \label{fig:carbon2}
  \end{minipage}
\end{figure*}

\end{document}